\pdfoutput=1

\documentclass[11pt,table,xcdraw,dvipsnames]{article}
\usepackage[preprint]{acl}

\usepackage{times}
\usepackage{latexsym}

\usepackage[T1]{fontenc}

\usepackage{microtype}

\usepackage{mdframed} 
\usepackage{booktabs}

\usepackage{inconsolata}
\usepackage{placeins}   

\usepackage{graphicx}
\title{ALIGN: Word Association Learning for Cultural Alignment in Large Language Models}

\author{Chunhua Liu\thanks{All authors contributed equally; author order is alphabetical by first name.}$^{1}$ \quad
        Kabir Manandhar Shrestha\footnotemark[1]$^{2}$ \quad
        Sukai Huang\footnotemark[1]$^{3}$ \\
        $^1$School of Computing and Information Systems, 
        The University of Melbourne\\
        $^2$Melbourne Data Analytics Platform, The University of Melbourne\\
        $^3$Faculty of Information Technology, Monash University\\
        \texttt{chunhua.liu1@unimelb.edu.au}\\
        \texttt{k.manandharshrestha@unimelb.edu.au}\\
        \texttt{sukai.huang@monash.edu}
}

\usepackage{CJKutf8}
\usepackage{anyfontsize}
\usepackage{multirow}
\usepackage{dashrule}
\usepackage{tabularray}
\usepackage{float}
\usepackage{tikz}
\usepackage{caption}
\usepackage{etoolbox}
\usepackage{makecell}
\AtBeginEnvironment{tblr}{\small}
\usepackage{enumitem}

\usepackage{algpseudocodex}
\usepackage{algorithm}
\usepackage{amsmath}
\usepackage{amssymb}
\usepackage[most]{tcolorbox}
\usepackage{newfloat}
\usepackage{soul}
\usepackage{listings}
\usepackage{wrapfig}

\usepackage{xspace} 
\newcommand{\swen}{\textsc{SWOW.en}\xspace}
\newcommand{\swzh}{\textsc{SWOW.zh}\xspace}
\newcommand{\swus}{\textsc{SWOW.us}\xspace}

\usepackage{subcaption}

\usepackage{listings}
\usepackage{xcolor}

\definecolor{codegreen}{rgb}{0,0.6,0}        
\definecolor{codegray}{rgb}{0.5,0.5,0.5}     
\definecolor{codepurple}{rgb}{0.58,0,0.82}   
\definecolor{backcolour}{rgb}{0.95,0.95,0.92}
\definecolor{yellow20}{HTML}{fddc69}
\definecolor{magenta30}{HTML}{ffafd2}
\newcommand{\highlight}[2]{\sethlcolor{#1}\hl{#2}}

\lstdefinestyle{arxivpy}{
  language=Python,
  frame=lines,                 
  framesep=4pt,                
  xleftmargin=1em,             
  framexleftmargin=1em,        
  numbers=left,
  numberstyle=\scriptsize\color{codegray},
  stepnumber=1,
  numbersep=8pt,
  basicstyle=\ttfamily\scriptsize,   
  backgroundcolor=\color{backcolour},
  showstringspaces=false,
  tabsize=2,
  keepspaces=true,
  breaklines=true,
  breakatwhitespace=true,      
  postbreak=\mbox{\(\hookrightarrow\)\space},
  keywordstyle=\color{codepurple},
  commentstyle=\color{codegray},
  stringstyle=\color{codegreen},
  aboveskip=1.0em,
  belowskip=1.0em
}

\usepackage{arydshln} 

\DeclareCaptionStyle{ruled}{labelfont=normalfont,labelsep=colon,strut=off}
\newcommand{\aref}[1]{Appendix~\ref{#1}} 
\floatstyle{ruled}
\newfloat{listing}{tb}{lst}{}
\floatname{listing}{Listing}

\begin{document}
\maketitle

\begin{abstract}
    Large language models (LLMs) exhibit cultural bias from over-represented viewpoints in training data, yet cultural alignment remains a challenge due to limited cultural knowledge and a lack of exploration into effective learning approaches. We introduce a cost-efficient and cognitively grounded method: fine-tuning LLMs on native speakers' word-association norms, leveraging cognitive psychology findings that such associations capture cultural knowledge. Using word association datasets from native speakers in the US (English) and China (Mandarin), we train Llama-3.1-8B and Qwen-2.5-7B via supervised fine-tuning and preference optimization. 
    We evaluate models' cultural alignment through a two-tier evaluation framework that spans lexical associations and cultural value alignment using the World Values Survey. Results show significant improvements in lexical alignment (16–20\% English, 43–165\% Mandarin on Precision@5) and high-level cultural value shifts. On a subset of 50 questions where US and Chinese respondents diverge most, fine-tuned Qwen nearly doubles its response alignment with Chinese values (13 $\rightarrow$ 25).
    Remarkably, our trained 7–8B models match or exceed vanilla 70B baselines, demonstrating that a few million of culture-grounded associations achieve value alignment without expensive retraining. Our work highlights both the promise and the need for future research grounded in human cognition in improving cultural alignment in AI models. 
\end{abstract}



\section{Introduction}
Every culture creates its own unique lens for understanding the world \citep{92fd11aa-f25a-3645-9c8e-e6bdcfb48adb}. While we all share the same basic human brain, the way we use it—how we think, feel, and make sense of reality—is fundamentally shaped by our cultural environment \citep{Cult_brain}. Through years of immersive experience, culturally specific ways of thinking become internalized \citep{Cult_point}. These deep mental frameworks automatically guide how we interpret concepts, perceive situations, and make decisions. At the same time, this long-term internalization makes cultural knowledge difficult to capture systematically. Much of this knowledge operates as common sense within a culture—deeply embedded and rarely articulated \citep{atlas_cult}. While some cultural information exists online, e.g., holidays and traditions, this represents only the visible surface \citep{hall1976beyond}. The deeper layers of cultural cognition, including unspoken assumptions, subtle social cues, and the implicit ways people naturally connect concepts, remain hidden within the minds of cultural insiders.

\begin{figure}[!t]
    \centering
    \includegraphics[width=\linewidth]{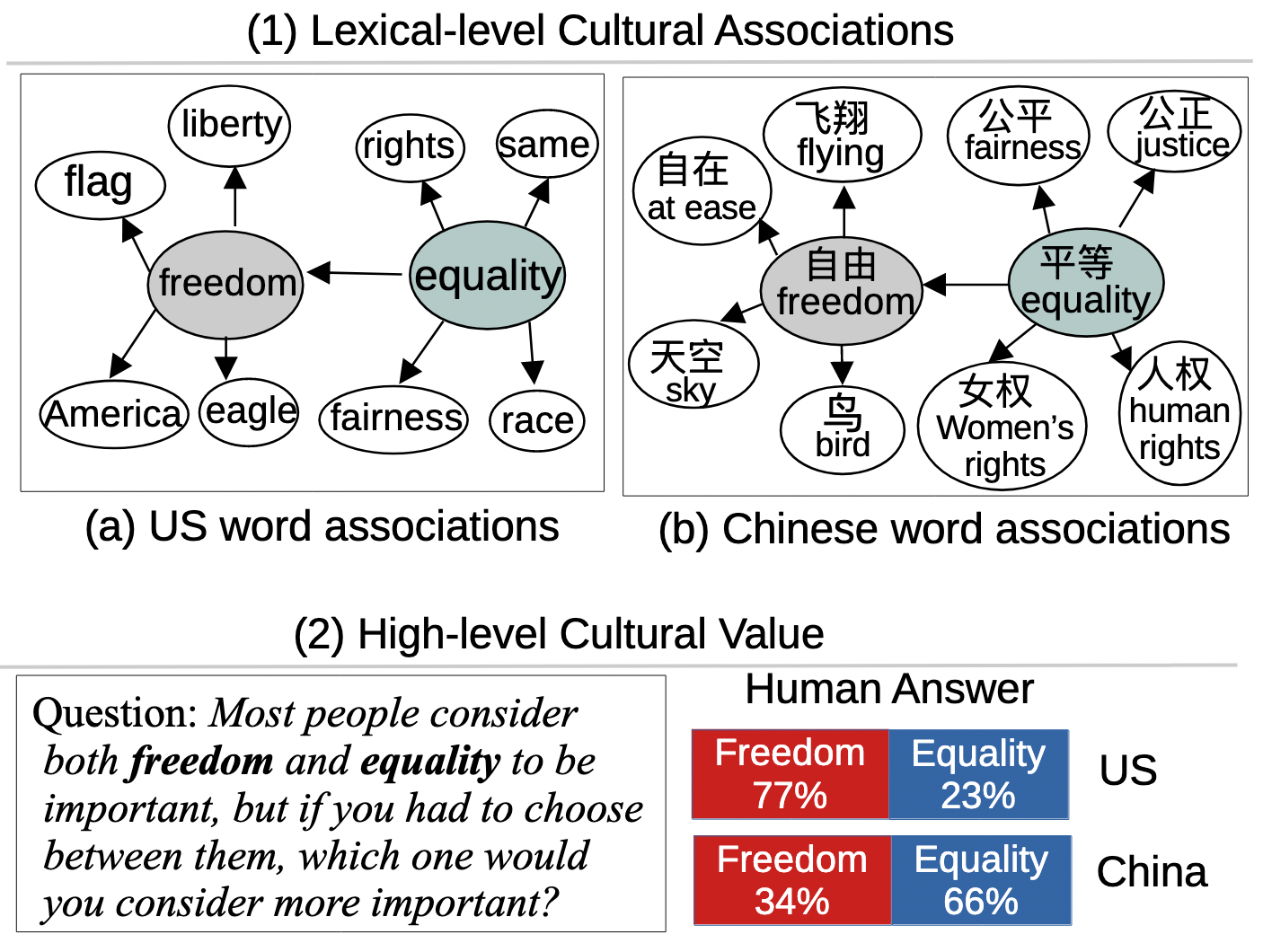}
    \caption{Example of how cultural word associations at the lexical level relate to higher-level cultural values. (1) Word associations show distinct cultural perception around the word of \textbf{freedom} and \textbf{equality}, with American associations emphasizing individual liberty and patriotic symbols, versus Chinese associations focusing on collective harmony and institutional frameworks. (2) These lexical differences correspond to opposing value preferences in responses to the survey question.}
    \label{fig:motivation_example}
\end{figure}

As large language models (LLMs) become embedded in global communication, they increasingly engage with users from diverse cultures. However, most LLMs are trained primarily on English-language data, leading to an over-representation of Western perspectives and an under-representation of cultural-specific concepts~\citep{cao-etal-2023-assessing, naous-etal-2024-beer}. This bias not only limits their effectiveness in culturally grounded applications \cite{Nguyen2024Mango}, but also risks ethical issues and inappropriate responses (e.g., suggesting drinking wine after Maghrib prayer~\cite{naous-etal-2024-beer}). Ensuring LLMs are culturally aware is crucial for fostering diversity and effective communication in today’s AI ecosystem~\cite{hershcovich-etal-2022-challenges}. Full retraining, however, is prohibitive: frontier models consume hundreds of petaFLOPs-days and tens of millions of dollars \cite{10.5555/3600270.3602446}, exacerbating carbon costs and the global ``AI compute divide'' \cite{faiz2024llmcarbon}. Parameter-efficient fine-tuning (LoRA, QLoRA) touches <1\% of weights and substantially reduces compute demands, yet still needs culture-rich data \cite{Hu2021LoRALA, 10.5555/3666122.3666563}. {Moreover, prior work~\cite{li2024culturellm} shows that training one universal model for all cultures is challenging 
and often less effective than culturally tailored models, 
as cultural knowledge can conflict or intertwine. We adopt the same view that culture-specific models are central to cultural alignment.}

Recent work has focused on evaluating cultural alignment using surveys \cite{durmus2024towards} and adapting models through prompting or synthetic data~\cite{cao2024cultural, shi-etal-2024-culturebank}, but without lived-experience corpora, true cultural grounding remains elusive \cite{Liu2025CultureTaxonomy}.

In response, we turn to native speakers’ free word associations—a classic psycholinguistic lens on implicit cultural representations. When prompted with \textit{red}, US respondents offer \textit{danger}, \textit{stop}, or \textit{anger}, whereas Chinese respondents give \textit{happiness}, \textit{celebration}, or \textit{luck}, illustrating how such spontaneous links reveal culture-specific representations. 
If such lexical links mirror deeper cultural values, aligning them should steer models toward cultural judgments. Figure~\ref{fig:motivation_example} provides an example of such transfer.

We use two training approaches to fine-tune Llama-3.1-8B and Qwen-2.5-7B on two word association datasets, English (\swus) and Mandarin (\swzh). Then we test (i) how well it regenerates human associations and (ii) World Values Survey alignment. Our findings reveal that (1) vanilla Llama initially leans more toward US associations and values than Qwen, whereas vanilla Qwen leans more toward Chinese associations and values than Llama; (2) association-tuned models produce markedly more human-like affective associations; and (3) this lexical gain translates into higher value alignment with the target culture, most notably when the original model lacked that knowledge. This work makes three key contributions: 
\begin{enumerate}
\item We present the first head-to-head study of cultural fine-tuning, contrasting LoRA-based supervised fine-tuning with preference-optimized models on the English and Chinese SWOW associations, demonstrating their potential as valuable cultural resources.
\item We show how lexical-level association training shifts models toward target-culture value judgments using a two-tier evaluation.
\item We will release\footnote{\scriptsize\url{https://github.com/acl-anon-2025/cultural-lexis-anon}} the training pipeline and the top-performing models, to support plugging US- or CN-specific adapters  into other LLMs and extending it to new cultures.
\end{enumerate}

\section{Related Work}



\subsection{Cultural Alignment in LLMs}

\textbf{Cultural Bias in LLMs} LLMs inherit the skew of their training corpora; the English-heavy web thus pushes models toward Western-centric values \cite{naous-etal-2024-beer, adilazuarda-etal-2024-towards}. In the absence of broad, authentic datasets, researchers mine cultural proxy sources such as Wikipedia \cite{nguyen2023extracting} and online communities \cite{shi-etal-2024-culturebank}, or ask LLMs to fabricate synthetic cultural data \cite{bhatia-shwartz-2023-gd, west-etal-2022-symbolic}. Yet, as \citet{Liu2025CultureTaxonomy} notes, lived-experience corpora remain scarce. We fill this gap by exploring large-scale native word-association norms as a direct, culturally grounded resource.

\textbf{Cultural Alignment Evaluation} Alignment is typically judged by comparing model outputs with human responses from specific cultures \cite{Liu2025CultureTaxonomy, adilazuarda-etal-2024-towards}. {These evaluations broadly fall into two categories: assessing cultural knowledge (e.g., food, customs) and evaluating high-level cultural values. Several recent benchmarks have been proposed for various cultural knowledge. However, these datasets are often either (a) domain-specific, e.g., FORK \citep{palta2023fork} only focuses on cutlery and food; (b) being verified/annotated by only a few (typically 2–5) native speakers, e.g., FORK was verified by two annotators, BLEnD~\cite{myung2024blend} and CulturalBench~\cite{chiu-etal-2025-culturalbench} were annotated by five annotators;
or (c) the questions being asked are predominantly English-centric (e.g., CulturalBench and FORK only include English).}
{On the value evaluation direction,} researchers draw on cross-national surveys such as Hofstede’s dimensions \cite{geert2020cultures} and the World Values Survey (WVS) \cite{haerpfer2020world}. These surveys are usually conducted within a large scale of native speakers within one country, and the resulting response often reflects the population level distribution. 
Recent benchmarks build on WVS to evaluate LLMs across nations (e.g., GlobalOpinionQA ~\cite{durmus2024towards},  WorldValueBench~\cite{zhao-etal-2024-worldvaluesbench} both provide English questions)
capitalizing on its large sample sizes and 200-country coverage. We likewise adopt WVS for our value-alignment tests in Section~\ref{sec:exp2_valuealign} {and extend it beyond English with Chinese, matching the native language of Chinese participants.} 


\subsection{Word Associations and Their Value}

{In a word association task, participants provide the first (three) responses that come to mind for a cue}, exposing the spontaneous links that structure semantic memory. Large normative datasets now exist: the University of South Florida norms \cite{nelson2004university} and the crowd-sourced Small-World-of-Words (SWOW) corpus, whose English version spans 12K cues and 3M responses \cite{de2019small}. Compared with distributional embeddings, human associations convey richer affective and multimodal information \cite{de2021visual}. Parallel SWOW collections in Dutch \cite{de2013better}, Spanish \cite{cabana2024small}, Chinese \cite{li2024large} and other languages provide language-specific resources that reflect culture directly in speakers’ lived experience.

\textbf{Word Association and Culture} Association norms already illuminate cultural contrasts: \textit{food} evokes cuisine-specific terms across groups \cite{guerrero2010perception, son2014understanding}, and \textit{health} links to \textit{wealth} in India but to \textit{sick} in the United States \cite{garimella-etal-2017-demographic}. Large SWOW corpora further identify culture-defining keywords in Spanish, Dutch, English and Chinese \cite{lim2024computational} and recover language-specific moral values \cite{ramezani2024moral}. However, whether such lexical-level signals can also steer LLMs toward higher-level value alignment remains open. We tackle this gap by fine-tuning models on cultural associations and testing their transfer to value judgments. While drafting this paper, we noticed a concurrent work~\cite{dai2025word} that also uses word associations to steer language models via linear transformations. Unlike their primary focus on culturally aware association generation, our work explores different learning approaches to scale and transfer.

\section{Framework Overview}

We aim to investigate the extent to which models trained on association-level cultural knowledge can transfer to higher-level value alignment. To this end, we train language models on language-specific human word associations\footnote{{As culture and language are closely intertwined, we approximate cultures by their primary spoken language~\cite{delanoy2020culture}}. We treat language-specific word associations as culturally grounded signals, reflecting the conceptual organization shaped by speakers' cultural experiences.} using two training strategies and two model families. We then assess each model on two tiers: (i) association generation and (ii) value alignment via survey questions. This section introduces data and training, while the evaluations are presented in Sections \ref{sec:exp1_wag} and \ref{sec:exp2_valuealign}.


\paragraph{Language and Culture Selection} We focus on English for US and Mandarin for China (CN) because they provide a clear cultural contrast.
These cultures differ in individualism vs. collectivism, emotional expression norms, and conceptual associations (as illustrated in  Figure~\ref{fig:motivation_example}). Additionally, both languages have large-scale, high-quality native speaker word association datasets available, making this a practically significant test case for cultural transfer learning. {While our study focuses on two cultures, the methodology can also be applied to others. Here, we focus on the mechanisms of training and evaluation framework.}

\paragraph{Word-Association Datasets}
\label{sec:word_assoicatiion_datasets}
We train on the largest \textit{Small‐World‐of‐Words} corpora: English SWOW (\swen; \citealp{de2019small}) and Mandarin SWOW (\swzh; \citealp{li2024large}).  
\swen\ (2011–2019) provides 12K cues and 3.6M responses from 90K native speakers in the United States (\ensuremath{\approx}50\%), United Kingdom, Canada, and Australia. Each cue was answered by 100 participants with three free associations. For our US analyses we retain only respondents whose country \emph{and} native language are United States, hereafter \swus.  
\swzh\ (2016–2023) comprises 10K cues and 2M responses from 40K Mainland Chinese speakers.  
Both \swus\ and \swzh\ are randomly split \textbf{by cue} into 80 \% train, 10 \% validation, and 10 \% test (used in Section~\ref{sec:exp1_wag} as the test set).




\paragraph{Model Selection} We choose two widely used model families as the subjects of our study to examine how language-specific word associations influence a model’s cultural behavior given its initial representations. Specifically, we use Llama3.1-8B-Instruct~\cite{grattafiori2024llama3herdmodels} and Qwen2.5-7B-Instruct~\cite{qwen2025technicalreport} as our baseline models and then fine-tune them on SWOW datasets.\footnote{Due to computational resource constraints, we limited our study to training models of 7/8B parameters.}{\footnote{{While our evaluation includes US and Chinese cultural datasets, we do not assume Llama or Qwen to be clean proxies for any national culture due to the multilingual and multicultural composition of their training data. Instead, we quantify each model’s initial alignment (Section 4–5) and investigate the relative shifts after fine-tuning on SWOW datasets.}}


\subsection{Training LLMs on Cultural Associations}
To investigate how models acquire culturally grounded knowledge from word associations, we leverage two signals from human association data: (a) what associations people produce and (b) their relative production frequencies. We employ one learning approach for each signal, described below.

\begin{CJK*}{UTF8}{gbsn}
\paragraph{Supervised Fine-tuning (SFT)} 
The first approach leverages the association lists themselves of a cue word, which capture how native speakers understand the cue. For example, for the cue word \textit{country}, English associations include \textit{nation}, \textit{state}, \textit{America}, and \textit{farm}. For its Chinese equivalent \textit{国家}, associations include \textit{中国} (China), \textit{人民} (people), \textit{国旗} (flag), and \textit{富强} (wealthy and powerful). We implement \emph{the word association generation task} in the SFT framework, training models to generate associations that are more aligned with human associations.\footnote{We provide more training details in {\aref{appendix:sft_prompt_instructions}.}} Given a training example $x = \langle c, \mathbf{w} \rangle$, where $c$ is a cue word and $\mathbf{w} = \langle w_1, w_2, \ldots, w_n \rangle$ is a list of associated words, the model is trained to generate $\mathbf{w}$ conditioned on the cue word $c$. The objective of SFT is to maximize the likelihood of the training data.\footnote{See \aref{apdx:sec:experiment_settings} for SFT hyperparameter setting details.}
\end{CJK*}

\begin{table*}[!t]
\centering
\begin{minipage}[t]{0.48\textwidth}
\centering
\footnotesize
\resizebox{\textwidth}{!}{%
\begin{tabular}{clccccc}
\toprule
Test &\textbf{$M$ Type} & \textbf{$M$ Class} & \textbf{Train$_\text{swow}$} & \textbf{P@5} & \textbf{P@10} &  \textbf{P@40}  \\ 
\midrule
\multirow{4}{*}{\rotatebox{90}{\swus}}&Vanilla & Llama & - & 0.754 & 0.609  & 0.295 \\
&SFT & Llama & US & \textbf{0.875}  & \textbf{0.773}   & \textbf{0.437}   \\
&Vanilla & Qwen & - & 0.633 & 0.502 & 0.238  \\
&SFT & Qwen & US & {0.761}  & {0.651}  & {0.327}   \\
\midrule
\multirow{4}{*}{\rotatebox{90}{\swzh}}&Vanilla & Llama & - & 0.260  & 0.181   & 0.057   \\
&SFT & Llama & ZH & \textbf{0.689}  & {0.556}   & {0.277}   \\
&Vanilla & Qwen & - & 0.481  & 0.364    & 0.159    \\
&SFT & Qwen & ZH & \textbf{0.689}  & \textbf{0.559} & \textbf{0.279}   \\
\bottomrule
\end{tabular}}
\caption{Word Association Generation Results.}
\label{tab:en_word_assoc}
\end{minipage}
\hfill
\begin{minipage}[t]{0.4\textwidth}
\centering
\footnotesize
\resizebox{\textwidth}{!}{%
\begin{tabular}{ccccc}
\toprule
Test &\textbf{$M$ Type}  & \textbf{$M$ Class}  & \textbf{Train$_\text{swow}$} & \textbf{Spearman $\rho$} \\
\midrule
\multirow{4}{*}{\rotatebox{90}{\swus}}& Vanilla & Llama &  - & 0.241 \\
& PPO  &  Llama &  US & 0.270 \\
& Vanilla &  Qwen & - & {0.292} \\
&  PPO  &  Qwen & US& \textbf{0.321} \\
\midrule
\multirow{4}{*}{\rotatebox{90}{\swzh}}&Vanilla & Llama &  - & 0.211 \\
& PPO & Llama &  ZH & 0.226 \\
& Vanilla & Qwen & - & {0.291} \\
& PPO & Qwen &  ZH& \textbf{0.323} \\
\bottomrule
\end{tabular}}
\caption{Word Association Ranking Results.}
\label{tab:en_ranking}
\end{minipage}
\end{table*}

\paragraph{Proximal Policy Optimization (PPO) Training} The second approach leverages the observation that some associations are produced more frequently than others in human word association data (e.g., \textit{nation} is more frequent than \textit{farm}). We use reinforcement learning with PPO \citep{schulman2017proximal} to train models to rank associated words according to their frequency, framing the task as a ranking problem. Given a cue word $c$ and its randomized associated words $\mathbf{w} = \langle w_1, w_2, \ldots, w_n \rangle$, the model predicts a list ranking $\mathbf{r^{\prime}} = \langle r^{\prime}_1, r^{\prime}_2, \ldots, r^{\prime}_n \rangle$ to match the ground-truth ranking $\mathbf{r} = \langle r_1, r_2, \ldots, r_n \rangle$, where $r_i$ indicates word $w_i$'s empirical rank based on human association frequency in SWOW. 
To reflect these human preferences, we use the Spearman rank correlation between $\mathbf{r^{\prime}}$ and  $\mathbf{r}$ to determine the reward. This reward signal guides policy updates via PPO, encouraging the model to produce association rankings that better align with human preferences.\footnote{Initially, we conducted preliminary experiments with multiple task formats to determine the most effective design for PPO training. See details in {\aref{appendix:ppo_task_formats}} and {\ref{apdx:sec:experiment_settings}}.}

\begin{table*}[ht]
  \footnotesize
  \centering
  \resizebox{\textwidth}{!}{%
    \begin{tabular}{c l c 
                    c 
                    >{\columncolor{orange!10}}c 
                    >{\columncolor{blue!10}}c 
                    c 
                    >{\columncolor{orange!10}}c 
                    >{\columncolor{blue!10}}c}
      \toprule
      \textbf{Test}&\textbf{Metric} &
      \textbf{Human} &
      \textbf{Llama\textsubscript{van}} &
      \textbf{Llama\textsubscript{ppo}} &
      \textbf{Llama\textsubscript{sft}} &
      \textbf{Qwen\textsubscript{van}} &
      \textbf{Qwen\textsubscript{ppo}} &
      \textbf{Qwen\textsubscript{sft}} \\
      \midrule
      \multirow{5}{*}{\rotatebox{90}{\swus}}  & Valence      & 5.514 & 5.398 & 5.403 & \textbf{5.543}\textsuperscript{*} & 5.337 & 5.352 & \textbf{5.484}\textsuperscript{*} \\
       & Arousal      & 4.244 & \textbf{4.272}\textsuperscript{*} & \textbf{4.238}\textsuperscript{*} & 4.214 & 4.192 & 4.183 & 4.192 \\
      & Concreteness & 3.644 & 3.378 & 3.355 & 3.582 & 3.368 & 3.349 & 3.535 \\
      & Emotional \% & 84.6\% & 78.2\% & 77.5\% & 75.5\% & 73.5\% & 73.5\% & 74.9\% \\
      & \%Conc | \%Abs | \%Unk &
      64.3/29.8/5.9 &
      52.8/37.9/9.3 &
      51.1/38.7/10.2 &
      56.8/29.0/14.2 &
      50.5/37.0/12.5 &
      50.4/37.2/12.4 &
      56.7/29.6/13.7 \\
      \midrule
      \multirow{5}{*}{\rotatebox{90}{\swzh}}& Valence      & 5.386 & 5.341 & 5.311 & \textbf{5.427}\textsuperscript{*} & 5.352 & 5.332 & \textbf{5.411}\textsuperscript{*} \\
     & Arousal      & 5.378 & 5.258 & 5.270 & \textbf{5.408}\textsuperscript{*} & 5.233 & 5.220 & \textbf{5.370}\textsuperscript{*} \\
     & Concreteness & 3.657 & 3.370 & 3.394 & 3.576                         & 3.391 & 3.412 & 3.516                         \\
     & Emotional \% & 53.3\% & 31.8\% & 33.8\% & 41.9\%                     & 42.3\% & 41.6\% & 47.9\%                     \\
      & \%Conc | \%Abs | \%Unk &
      35.9/15.8/48.3 &
      17.9/12.7/69.4 &
      19.3/13.2/67.5 &
      27.6/13.2/59.2 &
      24.1/16.6/59.3 &
      24.2/15.8/60.0 &
      30.4/15.9/53.8 \\
      \bottomrule
    \end{tabular}%
  }
  \caption{Emotion and concreteness scores on \swus (top) and \swzh (bottom). \textsuperscript{*}\,Bold indicates no significant difference from human medians (\(p\ge 0.05\)).}
  \label{tab:emotion_conc_en_complex_us_full}
\end{table*}

\section{Association-level Evaluation}
\label{sec:exp1_wag}

We test whether fine-tuning taught the models human-like word associations.  
To this end, we conduct two complementary evaluations: (a) \textbf{intrinsic} evaluation that measures the model performance on the task they are trained on, i.e., word association generation task for SFT models and ranking task for PPO models; and (b) \textbf{extrinsic} evaluation that focuses on the psychological attributes of generated word associations and measures to what extent they align with native speakers’ associations regarding the emotional intensity (valence/arousal) and concreteness of meaning.\footnote{This psychological attributes evaluation is inspired by a recent study~\cite{xiang2025comparing} where they found that the word associations generated by the vanilla model (Llama3.1-8B-Instruct) tend to be less emotional and concrete than humans in English word associations, revealing a gap. We extend the analysis to US and Chinese fine-tuned models.}\footnote{Prior study in cross-cultural study also shows that different cultural emotional connotations can be reflected in word associations~\cite{tham2020systematic}, e.g., ``green'' $\rightarrow$ ``envy'' in US (from phrase “green with envy” means jealous) and ``green'' $\rightarrow$ ``hat'' in China (from the saying “wearing a green hat”, symbolizing unfaithfulness).}
For the \textbf{evaluation set} in this section, we use \textbf{10\% of held-out testing cues and their associations}, i.e., 10\% of English data from \swus and 10\% of Chinese data from \swzh (see Section~\ref{sec:word_assoicatiion_datasets} for details).


\subsection{Intrinsic Evaluation}  
For each cue in the test set, we use the same prompt as the training stage to elicit the model associations. We use Precision@$K$ (overlap with human top-$K$) on the word association generation task and Spearman~$\rho$ against human frequency ranks on the ranking task by following prior work on word association evaluations~\cite{yao-etal-2022-wordties}.

Table~\ref{tab:en_word_assoc} shows the results on word association generation. Overall, all models achieve higher performance in English than in Chinese, reflecting their stronger English capability. On the Chinese test set, vanilla Qwen outperforms Llama, showing that Qwen has stronger Chinese capability. 
Models trained on \swus and \swzh achieved substantial gains, with SFT models improving P@5 by 16–20\% in English and 43–165\% in Chinese.
Table~\ref{tab:en_ranking} shows the results of the ranking task on PPO models, which exhibit similar trends of improvement but to a lesser degree.

\subsection{Extrinsic Evaluation}
\label{sec:psycho}




We examine three psychological attributes of associations: valence (pleasantness), arousal (emotional intensity), and concreteness (tangibility). Our \textbf{approach} is as follows: (1) for a cue $c$, we obtain its top-10 model-generated associations; (2) for each association, we look up its emotion and concreteness scores from existing norms 
(EN: \citet{warriner2013norms} for emotion, \citet{brysbaert2014concreteness} for concreteness; 
ZH: \citet{xu2022valence} for emotion, \citet{Xu_andLi2020concreteness} for concreteness); (3) we compute the median emotion/concreteness score of model-generated and human associations for $c$; and (4) we compare these medians across all cues.\footnote{We use the Wilcoxon signed-rank tests~\cite{wilcoxon1992individual} to examine if the two sets of median scores are significantly different or not. See detailed methodology in Appendix~\ref{apdx:sec:eval_psycholocial}.} 


Table~\ref{tab:emotion_conc_en_complex_us_full} presents the experimental results.\footnote{Violin plots in \aref{app:valence_arousal_violin} are provided to show a finer-grained view of the distributions of the three attributes.}  Overall, associations generated by SFT models 
exhibit a similar degree of valence (pleasantness) and arousal (emotional intensity) as human associations (e.g., when prompted with \textit{Halloween}, both humans and fine-tuned models evoke pleasant associations such as \textit{party} and \textit{holiday}).
Yet, a persistent gap remains in concreteness, with all model associations being more abstract than human associations (lower concreteness scores) in both languages. While SFT training increases concreteness by +0.20–0.21 from vanilla models, 
they remain 
below human medians. {For example, for the cue word \textit{emotions}, human associations include both abstract words such as \textit{feelings} (1.68 concreteness score) and \textit{sadness} (1.82), as well as more concrete ones like \textit{tears} (4.56). In contrast, model associations are dominated by abstract concepts, such as \textit{feelings} (1.68), \textit{empathy} (1.63) and \textit{love} (2.07).}\footnote{More concrete examples on Valence, Arousal and Concreteness are provided in Table~\ref{tab:examples_psychological_attributes} in \aref{apdx:ssec:example_psychological_attributes}.}
{These analyses highlight the advances achieved and the remaining challenges in aligning cultural conceptual representations with native speakers.}

\section{Cultural Value Alignment Evaluation}
\label{sec:exp2_valuealign}

Fine-tuning on language-specific word associations embeds lexical cultural patterns, but \emph{does this knowledge support higher-order reasoning about cultural values and beliefs?} Next, we evaluate this transfer using the World Values Survey (WVS). Successful transfer of association-driven cues to value-based scenarios would demonstrate deeper cultural understanding; failure would imply the need for explicit training on higher-level cultural reasoning tasks. We first measure how well models align with target-culture responses, then analyze prediction shifts on a curated ``tension-set'' of questions to probe fine-grained cultural differences.

\subsection{Experimental Setup} 
\paragraph{Dataset}
We evaluate cultural value alignment using the WVS~\cite{haerpfer2022world},  {which contains 290 questions systematically designed to cover twelve cultural topics and have surveyed 
native speakers of each country (wave7: 2,596 in US and 3,036 in China). 
The WVS provides two critical advantages that are not available in other datasets~\cite{palta2023fork, myung2024blend, cabana2024small}: (1) \emph{population-level} value distributions enabling reliable ground-truth for cultural value estimation, and (2) \emph{parallel questions in native languages} that reveal cultural differences in responses to the same set of questions, allowing us measure how the models align with different cultural values.} 
{We focus on the two cultures in our training data: the United States and China. From the 290 original questions, we removed demographic items (Q260–290) and retained only those asked in both countries, yielding 221 questions for evaluation. 
During evaluation, we use the language that is aligned with the target culture to prompt the language models (Chinese for both the WVS questionnaire and the models trained on \swzh; English for US).\footnote{We collected the English and Chinese WVS questionnaire from the  \href{https://www.worldvaluessurvey.org/WVSDocumentationWV7.jsp}{official website}. We also adopted the prompts that the WVS was presented to the participants.}}


\paragraph{Evaluation}
We use vllm with constrained sampling to generate answers. For a given question, we constrain the output tokens to be the symbols of the options {(e.g., 1,2,3,4)} and constrain the output token number to be 1. Then we take the token logprob across the specified options and re-normalize them to get the distribution of the answer options \cite{robinson2023leveraging}.
{We measure the alignment using the distance between human answer distribution $P = (P_1, P_2, \dots, P_n)$ and the model predicted probability distribution $Q = (Q_1, Q_2, \dots, Q_n)$.\footnote{{$P$ and $Q$ are two discrete probability distributions on the same $n$-point ordered support ($\sum_{i=1}^n P_i = \sum_{i=1}^n Q_i = 1$).}} 
} We use two distance metrics that are used separately in prior work~\cite{durmus2024towards,zhao-etal-2024-worldvaluesbench}: (a) \textbf{Jensen-Shannon distance (JSD)}, {which measures the distributions differences using category-wise probability divergence without considering any relationship between categories;\footnote{{
$\mathrm{JSD}(P, Q) 
= \sqrt{\frac{D(P \parallel M) + D(Q \parallel M)}{2}}$, where $M = \frac{1}{2}(P + Q)$. The $D(. \parallel .)$ is the KL-Divergence of two distributions, e.g., $D_{\mathrm{KL}}(P \parallel M) = \sum_{i=1}^{n} P_i \log \frac{P_i}{M_i}.$}}} and (b) \textbf{Earth Mover's distance (EMD)}, {which measures the accumulated cost of moving probability mass along the ordered scale.\footnote{{$EMD(P, Q) = \frac{1}{n}\sum_{i=1}^{n} \big| \delta_i \big|$, where $\delta_0 = 0$ and $\delta_{i+1} = \delta_i + P_i - Q_i, \quad i = 0, \dots, n-1$.}}}
{Both $0 \le \mathrm{JSD}(P, Q) \le 1$ and $0 \le \mathrm{EMD}(P, Q) \le 1$, closer to 0 means better alignment. JSD is more suitable for categorical distributions, whereas EMD is more appropriate for ordinal distributions as it considers the accumulated distance between options. As WVS answers include both types, we use both metrics.}
{We evaluate performance on the test set at two levels: (a) aggregate overall scores computed over the entire set, and (b) a finer breakdown measuring the percentage of questions with distances below different thresholds of JSD/EMD.}
\begin{table}[!t]
    \footnotesize 
    \centering
    \resizebox{0.48\textwidth}{!}{%
    \begin{tabular}{clp{1.2cm}cll}
    \toprule
    \textbf{Test}& \textbf{$M$ Type} & \textbf{$M$ Class} & Train$_{\text{swow}}$ & \textbf{JSD$\downarrow$} & \textbf{EMD$\downarrow$} \\
    \midrule
    
      \multirow{8}{*}{\rotatebox{90}{WVS.US}} &  \cellcolor{gray!20}Vanilla & \cellcolor{gray!20}Llama & \cellcolor{gray!20}- & \cellcolor{gray!20}0.324~ & \cellcolor{gray!20}0.102~ \\
      & SFT & Llama & US & 0.392 & 0.114 \\	
      & PPO &  Llama & US & \textbf{0.288}$^*$ & 0.092 \\ 
      & \cellcolor{gray!20}Vanilla & \cellcolor{gray!20}Qwen & \cellcolor{gray!20}- & \cellcolor{gray!20}0.388 & \cellcolor{gray!20}0.131\\
      &SFT & Qwen & US & 0.355$^*$ & 0.118 \\ 
      & PPO  & Qwen & US & 0.353$^*$ & 0.125\\ 
    \cmidrule{2-6} 
     & Vanilla & Llama3.1\_70b & - & 0.294$^*$ & 0.094 \\ 
     & Vanilla & Qwen2.5\_72b & - & 0.262$^*$ & 0.109$^*$ \\
    \midrule 
     \multirow{8}{*}{\rotatebox{90}{WVS.CN}} & \cellcolor{gray!20}Vanilla & \cellcolor{gray!20}Llama & \cellcolor{gray!20}- & \cellcolor{gray!20}0.459~ & \cellcolor{gray!20}0.152~ \\
     & SFT & Llama & ZH & 0.421$^*$ & 0.129$^*$ \\ 
     & PPO  & Llama & ZH &0.334$^*$ &	0.112$^*$\\
     & \cellcolor{gray!20}Vanilla & \cellcolor{gray!20}Qwen & \cellcolor{gray!20}- & \cellcolor{gray!20}{0.415} & \cellcolor{gray!20}0.139 \\
     & SFT & Qwen & ZH & \textbf{0.325}$^*$ & \textbf{0.100}$^*$ \\
      & PPO& Qwen & ZH & 0.374$^*$	& 0.123$^*$ \\ 
      
     \cmidrule(lr){2-6} 
     & Vanilla & Llama3.1\_70b & - & 0.333$^*$	 & 0.100$^*$ \\ 
     & Vanilla & Qwen2.5\_72b & - & 0.328$^*$ & 0.116$^*$ \\
    \bottomrule
    \end{tabular}
    }
    \caption{World Values Survey results on US and CN. $\downarrow$ indicates that lower is better (higher alignment). * indicates the improvement over Vanilla is significant ($p<0.05)$.Prompting language matches the survey language (EN for US, ZH for CN).\label{tab:wvs_overall_results}}
    
\end{table}


\paragraph{Approaches}
We use Vanilla models as our \textbf{baseline} to understand their \textit{initial alignment} towards a specific culture.
We apply the same prompts as the Vanilla models to our fine-tuned models to measure the extent of cultural value transfer. We also include two 70B-scale models for zero-shot prompting, which allows us to contextualize our results more broadly and estimate the potential upper bound that word associations can provide.

\begin{figure}[t]
    \centering
    \small 
        
    
    \begin{subfigure}{0.35\textwidth}
        \centering 
        \includegraphics[width=\textwidth]{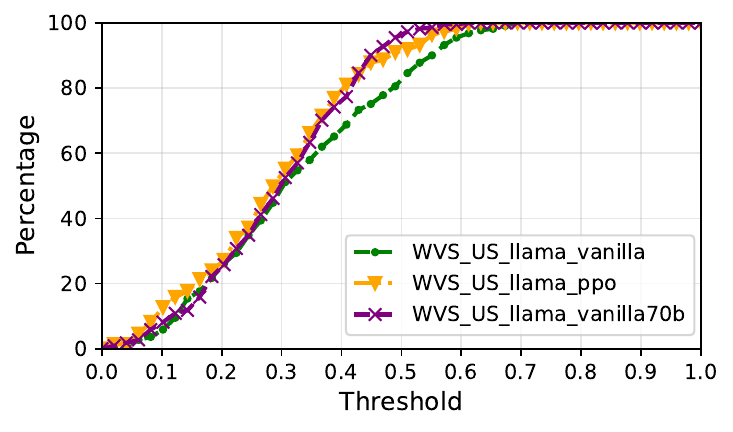}
        \caption{WVS.US performance under  JSD.}
        \label{fig:image4}
    \end{subfigure}
    \begin{subfigure}{0.35\textwidth}
        \centering
        \includegraphics[width=\textwidth]{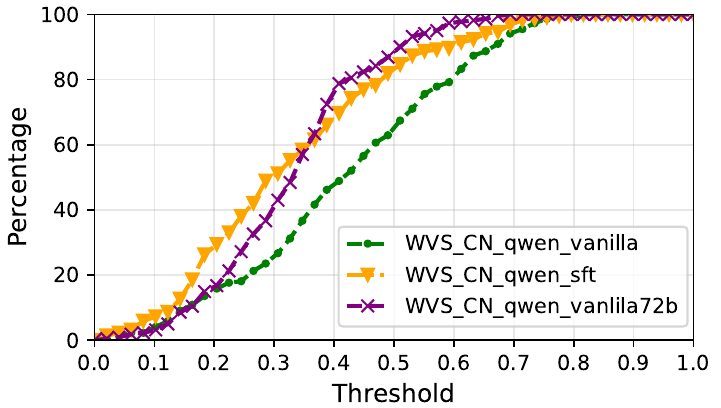}
        \caption{WVS.CN performance under JSD.}
        \label{fig:image3}
    \end{subfigure}
    \caption{Breakdown comparison of model alignment with cultural values across  United States (top) and China (bottom) based on the World Values Survey. Results are shown for the Vanilla and trained (SFT and PPO) versions of Qwen2.5 and Llama 3.1. The x-axis is the threshold for what counts as a ``good'' match, and the y-axis shows the percentage of questions where the model’s answer was within that threshold.}
    \label{fig:wvs_results_plots}
\end{figure}



\subsection{Overall Results}


Table~\ref{tab:wvs_overall_results} presents our results on WVS. Vanilla models exhibit different \emph{initial} degrees of cultural alignment with the target populations. 
In the \textbf{US setting (English)}, the Llama model shows better alignment with the ground-truth human responses compared to Qwen. While in the \textbf{CN setting (Chinese)}, the alignment trend reverses: the Qwen model outperforms Llama.
These findings align with our results in Section~\ref{sec:exp1_wag} and prior work that Llama models tend to be less eastern-value centric and less capable in understanding Chinese~\cite{xiang2025comparing, aksoy2025whose}, and Qwen has stronger Chinese capability~\cite{qwen2025technicalreport}. 

Models trained on the \textbf{\swzh} exhibit substantial improvements over their Vanilla models. 
Notably, \textbf{Qwen SFT model achieves the best performance on WVS.CN across the board.} 
Moreover, fine-tuned Llama aligns more closely with Chinese values, surpassing vanilla Qwen. 
This shows that \swzh provides strong cultural grounding and that training on it effectively steers the model toward high-level Chinese values. 

Training on \swus also brings significant improvements on WVS.US (except for SFT Llama). \textbf{The best-performing model is PPO Llama, which even achieves comparable or better results than the 70B models.} We also observe smaller overall gains on the US set than on the CN set,
suggesting that \swus might provide a weaker cultural signal than \swzh.
We hypothesize this may stem from models being highly exposed to English during pre-training compared to Chinese, or from the greater cultural diversity within the US increasing alignment difficulty.

\begin{figure}[!t]
  \centering
  \begin{subfigure}[b]{0.48\textwidth}
    \centering
    \includegraphics[width=\textwidth, height=.3\textheight, keepaspectratio]{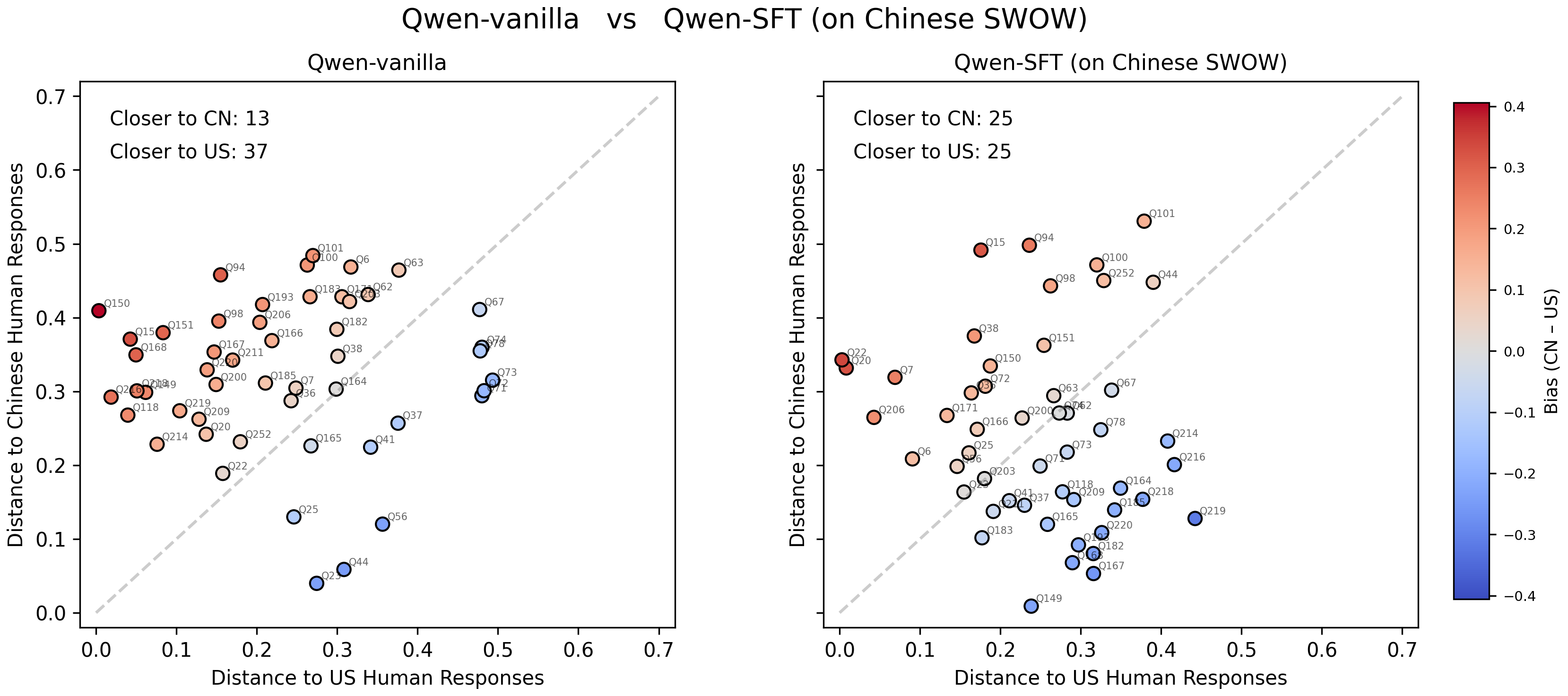}
    \caption{{Qwen2.5-7B}}
    \label{fig:qwen_shift}
  \end{subfigure}
  \begin{subfigure}[b]{0.48\textwidth}
    \centering
    \includegraphics[width=\textwidth, height=.3\textheight, keepaspectratio]{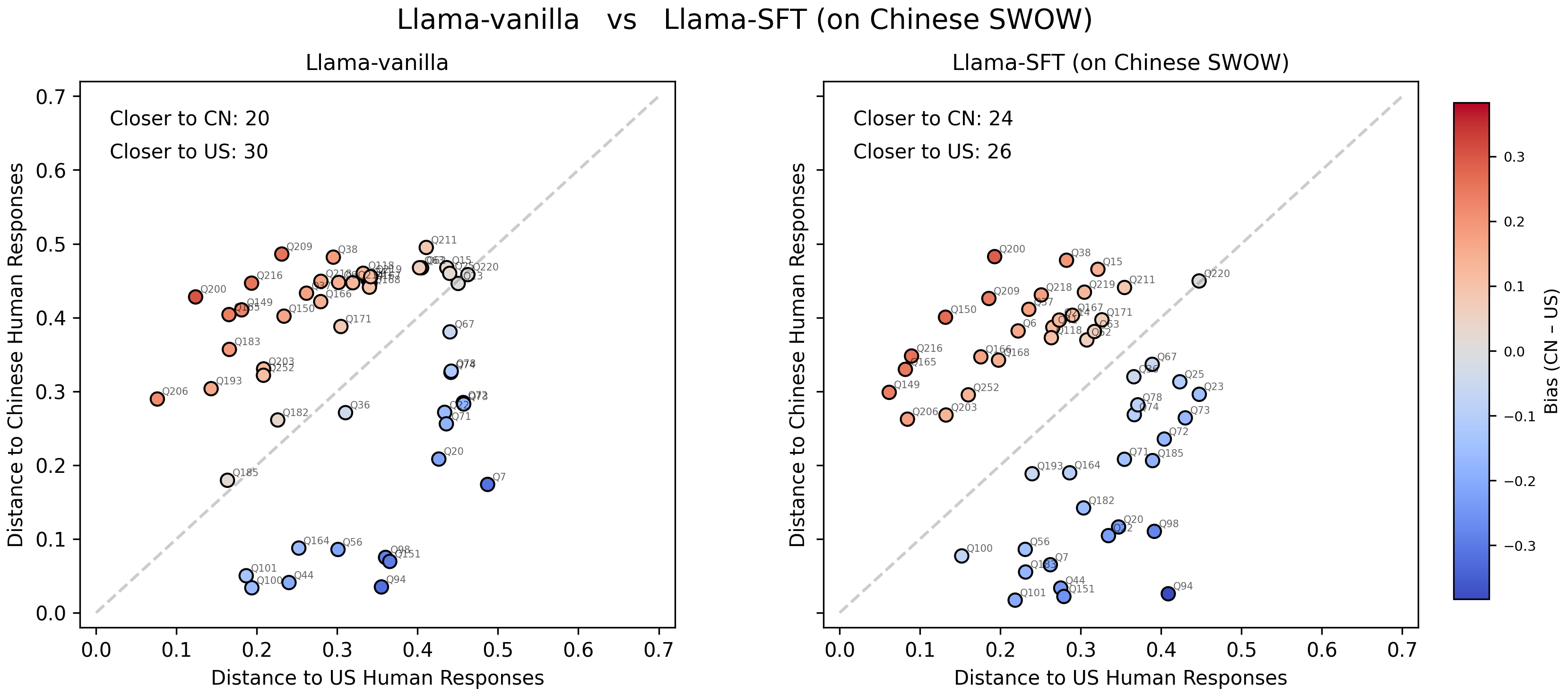}
    \caption{{Llama3.1-8B}}
    \label{fig:llama_shift}
  \end{subfigure}
  \caption{Comparison of shifts after SFT for Qwen and Llama on \swzh (ZH prompts). Each dot = one WVS question; blue (red) indicates that the question answer is more towards Chinese (English). Table~\ref{tab:tension_examples} presents concrete examples that illustrate the shifts.}
  \label{fig:shift_comparison}
\end{figure}

\begin{table*}[!th]
  {%
  \Huge            
  \renewcommand{\arraystretch}{1.6}  
  \centering
  \rowcolors{2}{gray!15}{white}
  \resizebox{\textwidth}{!}{%
    \begin{tabular}{l p{16cm} >{\centering\arraybackslash}p{7cm}>{\centering\arraybackslash}p{7cm} >{\centering\arraybackslash}p{7cm} >{\centering\arraybackslash}p{7cm} >{\centering\arraybackslash}p{7cm} >{\centering\arraybackslash}p{7cm}}
      \toprule
      
      \rowcolor{gray!30}\textbf{Id} & \textbf{WVS question (full wording + choice labels)} &
      \textbf{US} & \textbf{CN} &
      \textbf{Qwen\textsubscript{van}} & \cellcolor{blue!10}\textbf{Qwen\textsubscript{sft}} &
      \textbf{Llama\textsubscript{van}} & \cellcolor{blue!10}\textbf{Llama\textsubscript{sft}} \\
      \midrule
      Q149 &
        Most people consider both freedom and equality important, but if you had to choose between them, which would you consider more important? {\itshape\{1: Freedom; 2: Equality\}} &
        [77\%,23\%] & [34\%,66\%] &
        [83\%,17\%] & \cellcolor{blue!5}[33\%,67\%] &
        [93\%,7\%]  & \cellcolor{blue!5}[83\%,17\%] \\

      Q168 &
        In which of the following do you believe, if you believe in any? – Heaven {\itshape\{1: Yes; 2: No\}} &
        [65\%,35\%] & [12\%,88\%] &
        [71\%,29\%] & \cellcolor{blue!5}[18\%,82\%] &
        [97\%,3\%]  & \cellcolor{blue!5}[85\%,15\%] \\

      Q165 &
        In which of the following do you believe, if you believe in any? – God {\itshape\{1: Yes; 2: No\}} &
        [79\%,21\%] & [17\%,83\%] &
        [41\%,59\%] & \cellcolor{blue!5}[29\%,71\%] &
        [94\%,6\%]  & \cellcolor{blue!5}[87\%,13\%] \\

      Q118 &
        How often do ordinary people in your neighborhood have to pay a bribe, give a gift, or do a favor to local officials/service-providers to get needed services? {\itshape\{1: Never; 2: Rarely; 3: Frequently; 4: Always\}} &
        [28\%,55\%,15\%,2\%] & [4\%,34\%,36\%,26\%] &
        [33\%,55\%,10\%,2\%] & \cellcolor{blue!5}[5\%,19\%,67\%,10\%] &
        [93\%,4\%,2\%,1\%] & \cellcolor{blue!5}[77\%,9\%,8\%,6\%] \\

      Q166 &
        In which of the following do you believe, if you believe in any? – Life after death {\itshape\{1: Yes; 2: No\}} &
        [69\%,31\%] & [12\%,88\%] &
        [90\%,10\%] & \cellcolor{blue!5}[36\%,64\%] &
        [95\%,5\%]  & \cellcolor{blue!5}[87\%,13\%] \\
      \bottomrule
    \end{tabular}%
  }
  \caption{WVS questions where SFT on Chinese SWOW shifts Qwen’s distribution toward Chinese responses. Shaded cells highlight the fine-tuned model’s probabilities.}
  \label{tab:tension_examples}
  }
\end{table*}

Surprisingly, our best-performing trained 7/8B models even surpass some of the 70B models. For WVS.US, the 8B PPO-tuned Llama outperforms the Vanilla Qwen2.5 72B, while in WVS.CN, the 7B SFT-tuned Qwen outperforms the Vanilla Llama3.1 70B. Figure~\ref{fig:wvs_results_plots} further illustrates how well different models align with human responses, evaluated under varying thresholds of JSD.\footnote{{A similar trend on EMD is shown in~\aref{apdx:ssec:percentage_emd}.}} For both US and ZH settings, we include the best-performing fine-tuned model, its vanilla counterpart, and a larger model version. On WVS.US, the PPO-tuned model outperforms the vanilla model and even slightly surpasses the 70B model. On WVS.CN, the SFT model largely improves over the vanilla model across thresholds. For example, at a JSD threshold of 0.3, the SFT model achieves $\approx$50\% of the questions aligned, outperforming both the vanilla model (20\%) and the 72B model (40\%). These promising results highlight the potential of culturally grounded fine-tuning as a lightweight yet effective alternative to scaling up.






\subsection{Cross-Cultural Shifts}

Beyond assessing a model’s answers to a single culture, we track how the responses shift across cultures before and after fine-tuning on word-association data.
{Each model has its initial cultural leanings (e.g., Llama vanilla models align more closely with US values than Qwen, while Qwen aligns more with CN), so fine-tuning reveals both a model's adaptation to a target culture and its shift from its initial bias.}
We evaluate a model’s answers with respect to both US and China. To capture the shifts, we focus on WVS questions where Chinese and US participants' responses diverge strongly. We ranked divergence by the average of JSD and EMD, selecting the top 50 divergent questions.\footnote{See details on selecting the tension-set in {\aref{apdx:ssec:tension_set}.}}  
These ``high-tension'' questions provide greater sensitivity for detecting cultural shifts, allowing small changes in the model's alignment to become observable, whereas questions answered similarly by both populations offer little diagnostic value.


\paragraph{Results}
Figures~\ref{fig:qwen_shift} (Qwen2.5-7B) and
\ref{fig:llama_shift} (Llama3.1-8B) present the models’ prediction shifts before and after training in on \swzh.\footnote{{More results in US are provided in {\aref{appendix:ssec:en_prompt_alignment}}.}} For each of the 50 questions, we compare the model’s response distance to US answers (x-axis) against its distance to Chinese answers (y-axis). For \textbf{Qwen2.5-7B}, we find that Chinese-leaning responses increase from \textit{13} in the Vanilla model to \textit{25} after SFT, indicating a marked shift toward Chinese cultural preferences. For \textbf{Llama3.1-8B}, the Vanilla model's predictions are clustered along the diagonal and skewed toward the US, while the SFT-tuned Llama shifts more modestly, increasing from \textit{20} to \textit{24} Chinese-leaning responses, thereby reducing roughly one-third of its initial US bias. Table~\ref{tab:tension_examples} presents concrete `before-and-after' examples with human answer distributions (US, CN) and model prediction distributions, illustrating how SFT consistently shifts Qwen (and, to a lesser extent, Llama) away from the US majority proportions and toward the Chinese ones.

\section{Conclusion}
This study investigates how native speakers' word associations serve as cultural knowledge resources. We fine-tuned 8 language models (across two languages, two LLMs, and two training approaches) to learn cultural signals and evaluate their cultural alignment.
We find that fine-tuned mid-sized LLMs on language-specific word-association norms (English and Mandarin SWOW) yield clear improvements in both lexical and value alignment. Fine-tuned models retrieve human associations with higher precision and more closely match human valence and arousal ratings, while their World Values Survey responses shift toward target-culture distributions. 
These findings demonstrate that grounding LLMs in a few million associative cues can instill {authentic} cultural understanding and enhance value reasoning without costly retraining.


\section{Limitations}

\paragraph{Focusing on country-level alignment.} 
Our evaluation aggregates cultural values at the national level (United States vs. China) and does not employ persona- or demographic-based prompting. While this choice simplifies the analysis, it may mask important regional, social, or demographic variations within each country.

\paragraph{Temporal gap between data and model training.} 
We rely on WVS Wave 7 surveys conducted during 2017–2022 \citep{haerpfer2020world}, English SWOW associations collected in 2011–2018 \citep{de2019small}, and Mandarin SWOW data from 2016–2023 \citep{li2024large}. In contrast, Llama 3.1 (8B) and Qwen 2.5 (7B) were trained on web data up to late 2023/early 2024. This temporal mismatch means our human cultural benchmarks may not fully reflect the information learned by the models, and shifts in cultural values or associations after the data collection periods are not captured.

\paragraph{Limited scope of languages and models.} 
We focus on two high-resource languages (English and Mandarin) and two open-source models (Llama 3.1 and Qwen 2.5). 
This selection was chosen for tractability, but the findings might not generalize to other open-sourced model families or commercial models. Furthermore, the generalizability to low-resource languages and cultures requires further exploration. 
We consider cultural alignment research of using word associations as a two-step  program: (1) establish whether word associations can serve as transferable cultural knowledge, and (2) explore how this transfers to low-resource languages and determine minimal data requirements for effective transfer. Our study focuses on step (1), which provides the foundation for step (2). Given the positive results from our analysis and the open-sourced code, future work should extend to additional languages and model architectures. 

\paragraph{The Double-edged sword of cultural alignment.} In this study, we focus on the mechanism of learning cultural knowledge from the word associations dataset, and we observed closer cultural value alignment on the evaluated dataset. At the same time, we acknowledge that alignment might also bring risks, such as reinforcing stereotypes and amplifying existing biases. However, we argue that context-aware alignment mitigates harm more effectively than generic, one-size-fits-all models in contexts that require culturally specific adaptation. Universal models often default to Anglophone-majority norms; when user context differs, this mismatch is a common source of insensitive or inappropriate outputs. Our intent is to reflect cultural norms accurately within context, not to promote or endorse them. Accordingly, we recommend per-culture LoRA adapters used only in their intended context (e.g., ZH adapter for Chinese settings), with baseline fallback otherwise. While not part of this study, we suggest a deployment pattern that further mitigates harms: wrapping the per-culture adapters in lightweight agent safeguards (e.g., context routing/opt-in, a safety critic for stereotyping/generalizations, uncertainty-based abstention, and baseline fallback).

\paragraph{Impact of fine-tuning.} Our study evaluates how fine-tuning with culture-specific data directs the model toward a target culture. However, we do not evaluate the impact of fine-tuned models on non-targeted cultures. This is by design: our scope is not to train a universal model for all cultures, but to develop culture-specific models, as prior work shows that one-size-fits-all LLMs are inferior to cultural tailored models \citep{li2024culturellm}. In line with our stance on cultural alignment, as described in the Introduction, we regard culture-specific models as essential. Therefore, we train culture-specific models (e.g., via adapters, or specific checkpoints from PPO models) on a shared base model and evaluate them within the target culture. For languages without an available adapter (e.g., Dutch, German), we recommend using the baseline model rather than applying adapters tuned for other cultures (e.g., ZH-tuned or US-tuned adapters). Our released pipeline can also facilitate future work on expanding to more cultures.

\paragraph{Learning efficiency across cultures.} While our empirical results show that cultural association training improves alignment, they do not fully explain why certain learning approaches perform better under specific cultural contexts. For instance, SFT and PPO exhibit different learning efficiencies across cultures: SFT achieves optimal alignment with Chinese values, whereas PPO performs best on US values. These findings point to promising directions for future work to explore how training methods and cultural contexts interact, for example through analyses of model internal states.



\section*{Acknowledgements}
We thank Dr Lea Frermann and Dr Simon De Deyne for their 
insightful discussions and feedback. We also thank the reviewers for
their valuable comments.


\bibliography{main.bib}

\appendix

\raggedbottom                    
\setlength{\parskip}{0pt}       
\setlength{\parsep}{0pt}
\setlength{\topsep}{0pt}

\setlength{\textfloatsep}{4pt plus 1pt minus 1pt}  
\setlength{\floatsep}{4pt plus 1pt minus 1pt}      
\setlength{\intextsep}{4pt plus 1pt minus 1pt}     




\section{Fine-tuning LLMs on Cultural Associations}

Fine-tuning directly on word association  reshapes the model's behavior by adjusting its weight parameters. This approach has two key benefits:

\textbf{Independence from external KB:} \quad Fine-tuning eliminates the need for an external retrieval system during inference. RAG relies on real-time access to a knowledge base, which may not always be \emph{available} and can significantly slow down inference due to retrieval latency. In contrast, a fine-tuned model carries its learned associations internally, making it faster and more self-contained.

\textbf{Generalization beyond the dataset:}\quad Fine-tuning enables the model to generalize to unseen examples by learning patterns and semantic relationships during training. For example, since ``gorilla'' and ``monkey'' are close in the word embedding space due to their shared features, a model fine-tuned on ``monkey'' or other nearby words—whether as cue words or associations—can implicitly infer associations for ``gorilla'', even if it's absent from the dataset. 

In the following sections, we discuss the types of fine-tuning techniques and the associated task designs we employ for LLMs to learn word associations.

\subsection{Supervised Fine-tuning}

To provide context, we consider autoregressive LMs such as the GPT \citep{DBLP:conf/nips/BrownMRSKDNSSAA20} and Llama \citep{grattafiori2024llama3herdmodels} series, which generate tokens in a left-to-right, autoregressive manner. Let $\mathbf{x}{< i}$ be the first $i - 1$ tokens of a sequence $\mathbf{x}$, and let $x_i$ be the $i$-th token. The probability that the LLM predicts token $x_i$ at position $i$ can be written as $LM\theta(\hat x_i = x_i \mid \mathbf{x}{< i})$, where $LM\theta(\cdot)$ is the model's probability distribution over the vocabulary, and $\theta$ represents the model parameters.

We implement \emph{a word association prediction task} directly in the supervised fine-tuning (SFT) framework. Given a training example $x = \langle c, \mathbf{w} \rangle$, where $c$ is a cue word and $\mathbf{w} = \langle w_1, w_2, \ldots, w_n \rangle$ is a list of associated words, the model is trained to generate the associated words $\mathbf{w}$ conditioned on the cue word $c$. The objective of SFT is to maximize the likelihood of the training data, which is formalized as:

\begin{equation} 
    J(\theta) = \max_{\theta} \mathbb{E}_{\mathbf{x} \sim \mathcal{X}} \left[ \sum_{i=1}^{|\mathbf{x}|} \log LM_\theta(x_i \mid \mathbf{x}_{< i}) \right] 
\end{equation}

where $\mathcal{X}$ denotes the training dataset, and $|\mathbf{x}|$ is the length of the token sequence.

While this formulation captures the core learning objective, in practice we reformat each training instance into a more natural, instruction-style prompt that aligns with how LLMs are typically used. For example, we add constraints to the prompt to further guide the model's generation process, such as ``do not generate words conditioned on the presence of other words, but focus solely on the cue word.'' See \aref{appendix:sft_prompt_instructions} for details.

\subsection{PPO training}

To further align LLMs with culturally-informed word associations, we explore reinforcement learning from human feedback (RLHF), using Proximal Policy Optimization (PPO) algorithm \citep{schulman2017proximal}. RLHF has proven to be a powerful technique for fine-tuning LLMs by aligning them with preferences defined by a reward model, which is either trained on human feedback or based on predefined rules \citep{DBLP:conf/nips/Ouyang0JAWMZASR22, deepseekai2025deepseekr1incentivizingreasoningcapability}. Recent studies indicate that RLHF surpasses supervised fine-tuning (SFT) in enhancing LLMs' reasoning capabilities, as RLHF encourages exploration beyond explicit solutions found in training data, whereas SFT focuses on broad imitation of human-provided examples \citep{havrilla2024teaching, chu2025sft}. From an imitation-learning viewpoint, RLHF exhibits mode-seeking behavior, prioritizing precise modes of response distributions, which makes it particularly effective for reasoning tasks demanding accuracy \citep{xiao2025on}. For further details on the differences between these fine-tuning approaches, we refer readers to \citet{xiao2025on}.

We use a rule-based reward function designed to reflect the fulfillment of designed tasks. Before we turn into the task design, we first introduce the three components of RLHF framework: 
\begin{enumerate}
    \item a language model (policy) $LM_\theta$ generating candidate outputs,  
    \item a reward model $r(q, a)$ evaluating those outputs, where $q$ is the question and $a$ is the generated answer, and
    \item a reinforcement learning algorithm (e.g., PPO) that updates the model to maximize the received reward.
\end{enumerate}

Formally, RLHF fine-tunes the language model \( LM_\theta \) by optimizing the following objective:

\begin{flalign}
    &\max_{\theta} \mathbb{E}_{a \sim LM_\theta(a \mid q)} \left[ r(q, a) \right] \nonumber\\
    &- \beta D_{\text{KL}} \left[ LM_\theta(a \mid q)  ||  LM_{\text{ref}}(a \mid q) \right]
\end{flalign}
where $ LM_{\text{ref}}$ is a frozen reference model (typically the initial SFT model), and \( \beta \) is a scaling factor controlling the KL penalty that discourages large divergences from the reference model so as to maintain the model stability.

\textbf{Ranking-based format}\footnote{Initially, we conducted preliminary experiments with multiple task formats to determine the most effective design for PPO training. See details in {Appendix~\ref{appendix:ppo_task_formats}}.}\quad Ultimately, we settled on a ranking task, where the model was asked to rank a list of association words of a cue word based on its frequency in the SWOW dataset. This design offers a middle ground: (1) It is more structured and constrained than free-form generation, improving training stability and (2) It is more challenging than MCQ, providing useful reward gradients for learning.  

The reward function evaluates the alignment between the model's ranked list and ground truth rankings using Spearman's rank correlation coefficient. 

The objective of PPO is formalized as:

\begin{flalign}
    L_{\text{PPO}}(\theta) &= \mathbb{E}_{(c,\mathbf{w}) \sim \pi_\theta} \left[ \min\bigl(r(\mathbf{w})A, \right. \nonumber\\
    &\left. \mathrm{clip}(r(\mathbf{w}),1-\epsilon,1+\epsilon)A\bigr) - \beta\log q(\mathbf{w})\right]
\end{flalign}

\noindent where
\begin{flalign}
r(\mathbf{w}) &= \frac{\text{LM}_\theta(\mathbf{w}\mid c)}{\text{LM}_{\theta_{-1}}(\mathbf{w}\mid c)}, \\
q(\mathbf{w}) &= \frac{\text{LM}_\theta(\mathbf{w}\mid c)}{\text{LM}_{\mathrm{ref}}(\mathbf{w}\mid c)}, \\
A &= R_{\text{spearman}}(x) - V_{\text{critic}}(x)
\end{flalign}

While our main results focus on evaluating cultural alignment in downstream tasks, we also assess the LLMs’ performance on the training tasks themselves—namely, supervised fine-tuning (SFT) for word association prediction and PPO training for ranking tasks. These results provide hints into whether models have successfully learned word association patterns during fine-tuning. 


\subsection{Preliminary Experiments on Task Formats for PPO Training}
\label{appendix:ppo_task_formats}

One of the important preliminary experiments is to identify suitable task formats for PPO training, ensuring the complexity was balanced --- neither trivially solvable nor excessively challenging. Tasks that are too easy yield minimal gradients for learning, whereas excessively difficult tasks also prevent LLMs from exploring the correct answer.

We considered three task formats: Multiple Choice Questions (MCQ), Free-form Association Word Prediction, and Ranking-based Association Prediction. Below we discuss each format in detail along with our experimental findings.

\paragraph{Experiment 1: MCQ Format.}  
We initially designed an MCQ-style task to evaluate candidate answers consisting of different categories of word associations. Specifically, the model was presented with a cue word and required to choose the option (a set of associated words) most closely related to it. Each MCQ contained four categories of candidate answers:

\begin{itemize}
    \item Category 1: High-frequency direct associations
    \item Category 2: Low-frequency direct associations
    \item Category 3: Indirect associations (frequent associations of the cue’s frequent associations)
    \item Category 4: Random unrelated words
\end{itemize}

Table~\ref{tab:mcq_example} provides an illustrative example of this MCQ format.

\begin{table}[!t]
    \centering
    \small
    \begin{tabular}{lp{4cm}}
        \toprule
        \textbf{Category} & \textbf{Example Words (Cue: \emph{apple})}\\
        \midrule
        High-frequency & fruit, red, pear, tree \\
        Low-frequency & stem, sauce, farm, healthy\\
        Indirect association & internet, mouse, machine (from word \textit{computer})\\
        Random & house, planet, justice, notebook\\
        \bottomrule
    \end{tabular}
    \caption{An example illustrating MCQ task categories.}
    \label{tab:mcq_example}
\end{table}

We hypothesized that Category 2 (low-frequency direct associations) and Category 3 (indirect associations) would serve as hard negative distractors, enhancing task difficulty. However, our experiments revealed that Vanilla LLMs were able to solve these MCQs easily, achieving accuracy consistently near 100\%. Thus, we concluded that the MCQ format was too simplistic to generate meaningful reward gradients for PPO training.

\paragraph{Experiment 2: Free-form Word Prediction.}
Our next experiment involved training PPO directly on the original word-association prediction task used for supervised fine-tuning (SFT). Here, the model freely generated association words conditioned solely on the cue word without explicit constraints.

This task proved to be overly challenging. The space of potential actions and states was extremely large, causing PPO training to suffer from poor convergence. The model rarely explored words sufficiently close to the ground-truth associations, leading to sparse reward signals, which hindered effective training.

\paragraph{Final Selection: Ranking-based Format.}  
Ultimately, we selected a ranking-based format (as described in the main text), where the model ranks a provided list of association words for each cue word, ordered by their frequency in the SWOW dataset. This task strikes a suitable balance between structured guidance (to avoid sparse reward signals) and sufficient complexity (to prevent trivial performance), enabling effective gradient signals to guide PPO optimization.

\section{Prompts for Supervised Fine-tuning}
\label{appendix:sft_prompt_instructions}

We reformat each training instance into a more natural, instruction-style prompt that aligns with how LLMs are typically used. Below is a sample prompt for the cue word ``mosquito'' and its associated words:

\begin{tcolorbox}[title={Supervised Fine-tuning Example for English SWOW word association prediction},
    breakable,
    enhanced,
    fontupper=\small\small\ttfamily\fontfamily{cmtt}\selectfont,
    fontlower=\small\ttfamily\fontfamily{cmtt}\selectfont
]

\highlight{yellow20}{[CONTEXT]}

You are a sophisticated language model designed to explore word associations comprehensively.

Given a cue word, your task is to generate a comprehensive list of words associated with the cue word. Aim to cover as many relevant contexts, uses, and meanings as possible without repeating similar concepts. List a target of \highlight{magenta30}{[LOWER BOUND SIZE]} to \highlight{magenta30}{[UPPER BOUND SIZE]} words that together provide a broad and insightful representation of all significant associations. Focus on revealing both common and unique aspects related to the cue word to ensure a balanced and thorough exploration of potential associations. Words should be distinct from each other. Your response shall only be the list of associated words. Do not generate words conditioned on the presence of other words but rather focus on the cue word itself.

\highlight{yellow20}{[CUE WORD]}

mosquito

\highlight{yellow20}{[ASSOCIATED WORDS]}

bite, bug, itch, buzz, malaria, insect, blood, net, fly, annoying, pest, summer, ouch, itchy, buzzing, repellent, small, swat, irritating, gnat, netting, camping, midge, proboscis, river, pain, lump, sting, flight, disease, spray, slap, swamp, fever, allergy, annoyance, worthless, nest, crunchy, smack, huge  in  canada, dead, amazonian, insect  bite, awake, tropical, water, female, anopheles, coast, valentine, doug, tent, jungle, whine, bumblebee, bored, nozzle, blood  sucker, noisy, nasty, skin, vampire, torment, hawk, ear, itchy  welt, pinch, needle, dengue, africa, bloodsucker, annoying  bug, mosquito  net, australia, horrible, kill, ugly, genetics

\end{tcolorbox}

\begin{tcolorbox}[title={Supervised Fine-tuning Example for Mandarin SWOW word association prediction},
    breakable,
    enhanced,
    fontupper=\small\small\ttfamily\fontfamily{cmtt}\selectfont,
    fontlower=\small\ttfamily\fontfamily{cmtt}\selectfont
]
\highlight{yellow20}{[CONTEXT]}
\begin{CJK*}{UTF8}{gbsn}

您是一款专为全面探索词语关联而设计的高级语言模型。
给定一个提示词，你的任务是生成一个与该提示词相关联的全面词汇列表。目标是尽可能涵盖所有相关的语境、用法和含义，避免重复相似的概念。列出目标数量为 \highlight{magenta30}{[LOWER BOUND SIZE]} 到 \highlight{magenta30}{[UPPER BOUND SIZE]}个词，这些词共同提供对所有重要关联的广泛而深刻的表示。专注于揭示与提示词相关的常见和独特的方面，以确保对潜在关联进行平衡而彻底的探索。词语应彼此不同。你的回答只能是相关联的词语列表。不要生成受其他词语存在影响的词语，而是专注于提示词本身。
\end{CJK*}

\highlight{yellow20}{[CUE WORD]}

\begin{CJK*}{UTF8}{gbsn}
狱警
\end{CJK*}

\highlight{yellow20}{[ASSOCIATED WORDS]}

\begin{CJK*}{UTF8}{gbsn}
监狱， 警察， 警棍， 囚犯， 制服， 罪犯， 犯人， 凶， 看守， 坐牢， 严厉， 警犬， 暴力， 很凶， 手铐， 监管， 刑警， 局长， 公安， 强悍， 抹布， 铁窗泪， 打架， 叮当作响， 囚服， 斯雷因， 管理， 刑罚， 敬业， 可怕， 辛苦， 工作， 黑暗， 霸王， 钥匙， 牢饭， SM， 冷漠， 凶恶， 逃狱， 逃跑， 强壮， 酷刑， 狱都市变， 坏人， 凶悍， 男人， 刑法， 条纹服， 黑猫警长， 铁牢， 卓别林， 狱卒， 反派， 美剧， 狱中杂记， 法律， 僻静， 虐待， 劳改， 悔恨， 棍棒， 牢房， 殴打， 性虐待， 女警， 典狱长， 警装， 严格， 帅哥， 肉文， 铁棍， 警服， 电网， 高墙， 严肃， 警司， 很辛苦， 害怕， 抓人， 阳光， 美国， 斯坦福大学， 越狱
\end{CJK*}
\end{tcolorbox}

To prevent overfitting and pattern memorization during training, we randomly set the lower and upper bounds for the number of associated words required in each training instance. The associated words are not shuffled; instead, they are ordered by frequency from the SWOW dataset, with the most frequent words listed first. This ordering introduces an inductive bias, encouraging the model to think of the most common associations first.

\section{Prompts for PPO training}
\label{appendix:ppo_prompt_instructions}





The task for PPO training is to rank a list of association words of a cue word based on its frequency in the SWOW dataset. The prompt for PPO training is similar to that of SFT, but with a different instruction.

\begin{tcolorbox}[title={PPO training Example for English SWOW ranking task},
    breakable,
    enhanced,
    fontupper=\small\small\ttfamily\fontfamily{cmtt}\selectfont,
    fontlower=\small\ttfamily\fontfamily{cmtt}\selectfont
]

\highlight{yellow20}{[CONTEXT]}

You are a sophisticated language model designed to explore word associations comprehensively.

Given the cue word, rank the following associated words from the most strongly related (rank 1) to the least strongly related (rank 10). 

Important Notes: 1. Rank ONLY the provided associated words from strongest (1) to weakest (10) in relation to the cue word. 2. Do NOT introduce any new words that aren't in the provided list.

Think step by step, comparing each associated word to the others to determine their relative strength of association with the cue word.

**Your final answer should at the end of the response and be in the following format:**

Final Ranking:
Rank 1: [Associated Word]
Rank 2: [Associated Word]
...
Rank 10: [Associated Word]

\highlight{yellow20}{[CUE WORD]}

dislike

\highlight{yellow20}{[TARGET ANSWER]}

Rank 1: detest

Rank 2: orange

Rank 3: flavor

Rank 4: displeasure

Rank 5: be well

Rank 6: kid refusing to eat

Rank 7: ugh

Rank 8: boss

Rank 9: peeve

Rank 10: gas

\end{tcolorbox}

\section{PPO Reward function details}
\label{appendix:ppo_reward_details}

\begin{lstlisting}[style=arxivpy]
% def compute_reward(queries, prompts, labels):
     """
     Computes reward scores for PPO training based on Spearman's rank correlation
     between predicted and ground-truth word association rankings.
   
     Args:
         queries: List of model responses (each includes both prompt and response).
         prompts: List of prompt texts.
         labels: List of ground-truth ranked word lists.

     Returns:
         A tensor of Spearman correlation scores, one per example.
     """
     rewards = []
     for query, prompt, label in zip(queries, prompts, labels):
         # Extract the response by removing the prompt part
         response = query[len(prompt) - 1:]

         # Parse predicted rankings (e.g., "1: cat, 2: dog, ...")
         predicted_words = parse_ranked_words(response)

         # Normalize and filter ground truth
         ground_truth = [w.lower() for w in eval(label)]
         predicted_filtered = [w for w in predicted_words if w.lower() in ground_truth]

         # Convert to rank indices
         pred_ranks, gt_ranks = map_to_rank_indices(predicted_filtered, ground_truth)

         # Compute Spearman correlation
         score = spearmanr(pred_ranks, gt_ranks).correlation
         rewards.append(score if not pd.isnull(score) else -1.0)

     return torch.tensor(rewards, dtype=torch.float32)
\end{lstlisting}

\section{Experiment Settings}
\label{apdx:sec:experiment_settings}

The experiments were conducted using two compute nodes equipped with 4 NVIDIA A100 GPUs per node. For SFT, we used Llama Factory library. The hyperparameters are provided in Table \ref{table:sft_hyperparameters}.

\begin{table}[htbp]
    \centering
    \small
    \begin{tabular}{ll}
        \toprule
        \textbf{Hyperparameters} & \textbf{Value}\\
        \midrule
        Fine-tuning method & LoRA \\
        LoRA Rank &      64\\
        LoRA Alpha            & 256 \\                  
        Learning rate         & 1.0e-5   \\         
        Scheduler             & Cosine (warmup ratio=0.1)\\
        Batch size per GPU    & 18         \\  
        Gradient accumulation & 2    \\             
        Number of epochs      & 1.5     \\   
        Precision             & bf16     \\              
        Max sequence length   & 2048     \\         
        \bottomrule
    \end{tabular}
    \caption{Hyperparameters for SFT Training}
    \label{table:sft_hyperparameters}
\end{table}

For PPO training, we used OpenRLHF library. The hyperparameters are provided in Table \ref{table:ppo_hyperparameters}.

\begin{table}[!t]
    \centering
    \small
    \begin{tabular}{ll}
        \toprule
        \textbf{Hyperparameters} & \textbf{Value}\\
        \midrule
        Actor learning rate      & 5e-7                  \\
        Critic learning rate     & 9e-6                  \\
        Initial KL coefficient   & 0.1                   \\
        Micro train batch size   & 8                     \\
        Train batch size         & 32                    \\
        Micro rollout batch size & 16                    \\
        Rollout batch size       & 64                    \\
        Max training samples     & 1,000,000             \\
        Max epochs               & 1                     \\
        Prompt max length        & 1024                  \\
        Generation max length    & 1024                  \\
        Zero optimization stage  & 3                     \\
        Precision                & bf16                  \\
        Gradient checkpointing   & Enabled               \\
        Optimizer offload        & Adam offload          \\
        Attention implementation & Flash attention       \\
        VLLM tensor parallel size & 2                     \\
        \bottomrule
    \end{tabular}
    \caption{Hyperparameters for PPO Training}
    \label{table:ppo_hyperparameters}
\end{table}

\section{Evaluation on the Emotions and Concreteness}
\label{apdx:sec:eval_psycholocial}

\subsection{Psychological Norms}
For English, we evaluate the emotions in associations using the Valence, Arousal, Dominance (VAD) dataset \cite{warriner2013norms} with 13,915 English lemmas. 
A score close to 1 suggests that the concept tends to evoke a relaxed, bored, or sleepy emotional state, indicating a low arousal response, whereas a score near 8 signifies that the concept tends to be associated with feelings of excitement, happiness, or high arousal. Concreteness score is obtained from a lexicon with 40K English word lemmas~\cite{brysbaert2014concreteness}. Highly concrete concepts (a score within the range of 4 to 5) are defined as those that can be directly experienced through the senses, such as objects, actions, or sensations that are easily experienced. 

For Chinese, we use a lexicon with 11K simplified Chinese words for the Valence and Arousal \cite{xu2022valence}. 
For valence ratings, each word is rated on a seven-point scale: “–3” = extremely negative, “0” = neutral, and “+3” = extremely positive. For arousal ratings, each word is rated on a five-point scale: “0” = very low arousal and “4” = very high arousal. For concreteness in Chinese, we use a lexicon of 9877 Two Character Chinese words \cite{Xu_andLi2020concreteness}. Each word is mapped into a 1 to 5 score, where ``1'' = ``very concrete'' and ``5'' = ``very abstract''.

\subsubsection*{Pre-processing}
\begin{itemize}
    \item \textbf{Token cleaning}: d-case, strip punctuation; English
          tokens are WordNet-lemmatised using NLTK~\cite{bird2006nltk}, while Mandarin tokens remain in
          surface form after Chinese punctuation removal.
    \item \textbf{Lexicon look-up}: tokens are matched against the
          English VAD norms\,\citep{warriner2013norms} and concreteness norms
          \citep{brysbaert2014concreteness}, or the corresponding Mandarin lexicons
          \citep{xu2022valence, Xu_andLi2020concreteness}.  Tokens absent from a lexicon
          are ignored for that metric.
\end{itemize}

\subsubsection*{Hypothesis testing}
Cue‐level medians are compared with a paired Wilcoxon signed‐rank test to determine whether the model’s lexical profile is
\emph{indistinguishable} from that of humans.

We test whether a model’s typical score is \emph{statistically
indistinguishable} from the human baseline, so the null states
“no difference’’ while the alternative states “some
difference’’.\\ \\ 
\textbf{\emph{Null hypothesis} \(H_0{:}\;
\tilde{x}_{\text{model}} = \tilde{x}_{\text{human}}\)
(assumes equality).\\}
\textbf{\emph{Alternative} \(H_1{:}\;
\tilde{x}_{\text{model}} \neq \tilde{x}_{\text{human}}\)
(assumes a non-zero gap).\\ \\}
Cells with \(p \ge 0.05\) (i.e.\ we \emph{fail to reject}
\(H_0\)) are highlighted in \textbf{bold}.

\subsection{Cue-level Valence, Arousal and Concreteness}
\label{app:valence_arousal_violin}

Section~\ref{sec:exp1_wag} (Table~\ref{tab:emotion_conc_en_complex_us_full}) presents the median scores for the psychological attribute evaluation. 
Since the median tends to compress information, we further visualize the distributions of these scores across all cues to better illustrate variations in each attribute. 
Figures~\ref{fig:violin_plot_concreteness} (Concreteness), 
\ref{fig:violin_plot_arousal} (Arousal), and 
\ref{fig:violin_plot_valence} (Valence) 
show the distributions of the psychological attributes.

In these violin plots, we can clearly see that models fine-tuned on association datasets tend to exhibit distribution shapes more similar to those of humans (shown on the far left). 
For example, in English, the concreteness scores (Figure~\ref{fig:violin_plot_concreteness}) of both SFT models display a noticeable bulge in the upper range—resembling the human distribution—whereas the vanilla models show a more evenly dispersed pattern.

\begin{figure}[!t]
    \centering
    \captionsetup{font=small}
    \begin{subfigure}{0.46\textwidth}
        \centering
        \includegraphics[width=\textwidth]{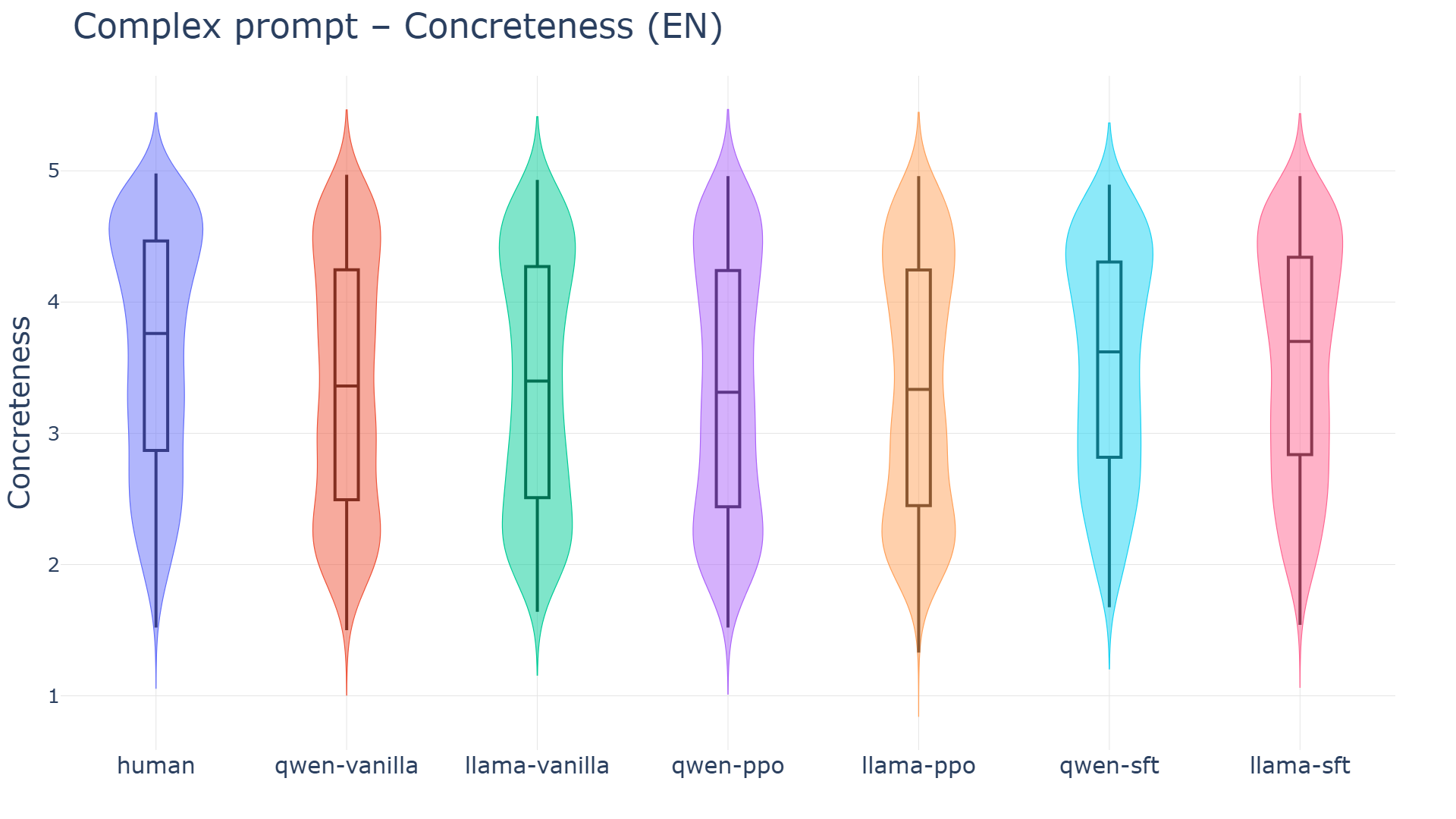}
        \caption{English: association \textbf{concreteness} (1 = abstract, 5 = concrete).}
        \label{fig:image1}
        
    \end{subfigure}
    \hfill
    \begin{subfigure}{0.46\textwidth}
        \centering
        \includegraphics[width=\textwidth]{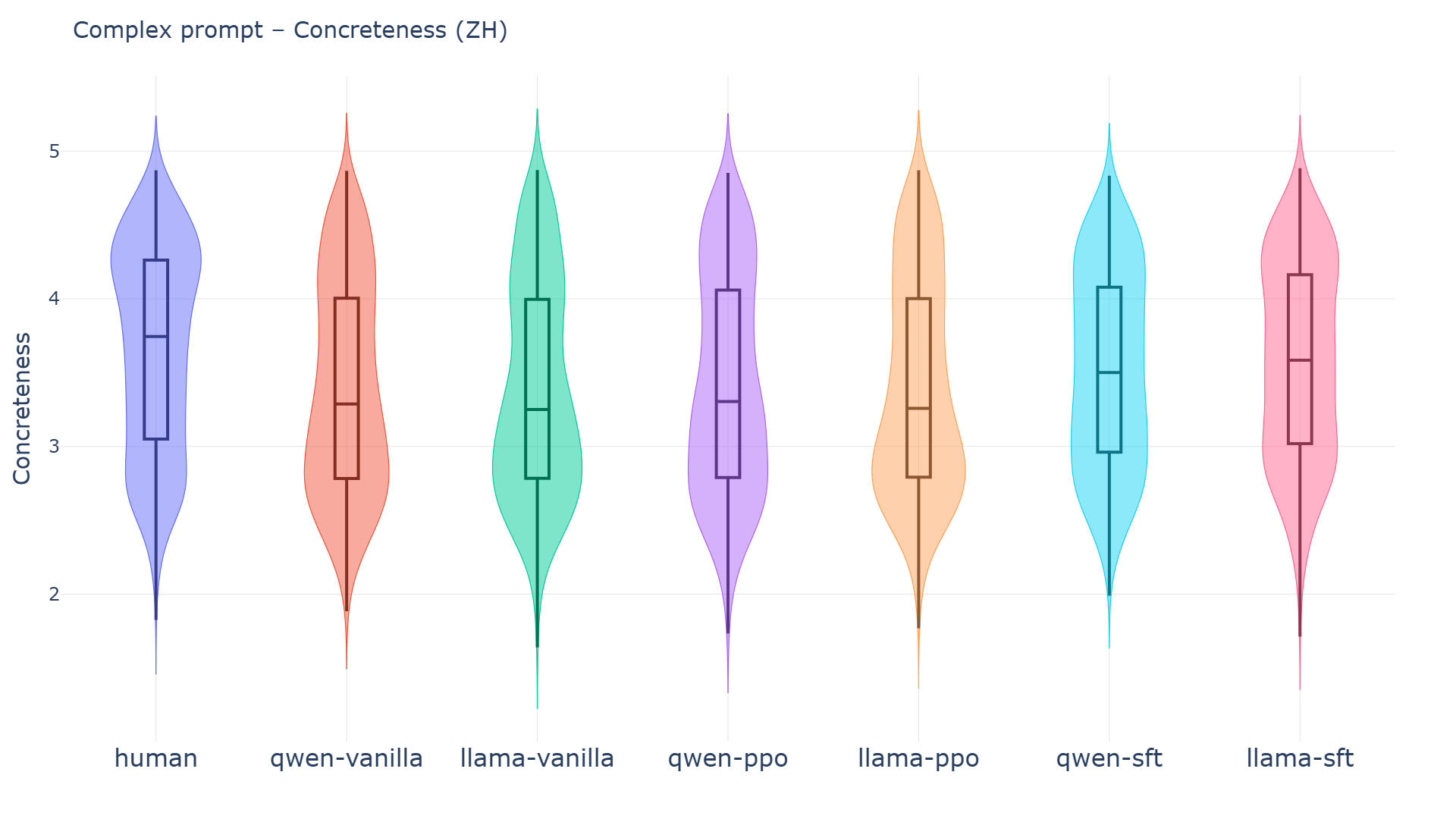}
        \caption{Mandarin: association \textbf{concreteness} on the rescaled 1–5 range.}
        \label{fig:image2}
    \end{subfigure}
    \caption{Violin + box plots of per-cue \textbf{concreteness} medians for the \emph{Complex} prompt. Left: English (1 = abstract, 5 = concrete); Right: Mandarin (rescaled to 1–5).}
    \label{fig:violin_plot_concreteness}
\end{figure}

\begin{figure}[!t]
\centering
    \captionsetup{font=small}
    \begin{subfigure}{0.46\textwidth}
        \centering
        \includegraphics[width=\textwidth]{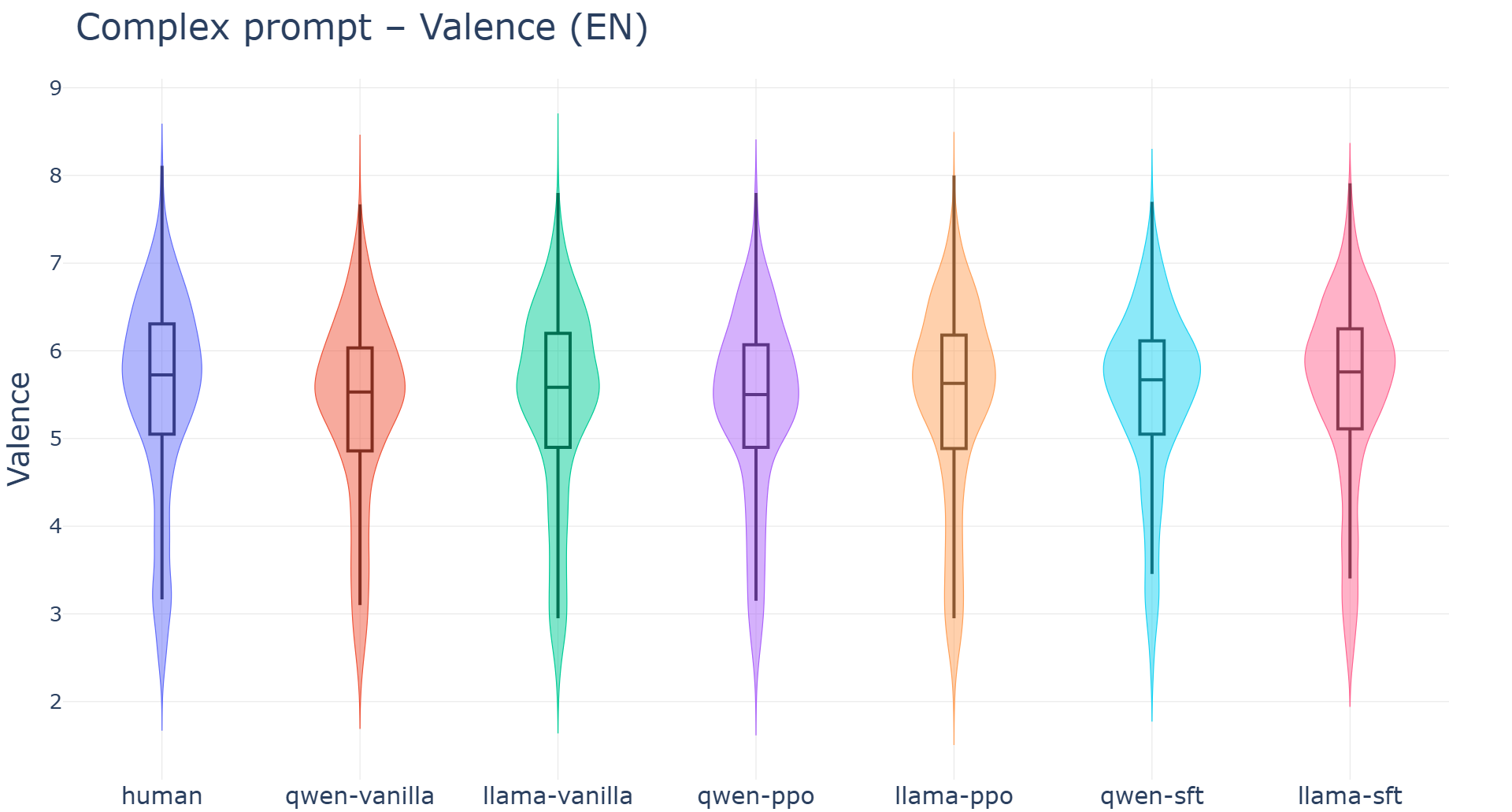}
        \caption{English: association \textbf{valence} (1 = unpleasant, 9 = pleasant).}
        \label{fig:valence_en}
    \end{subfigure}
    \hfill
    \begin{subfigure}{0.46\textwidth}
        \centering
        \includegraphics[width=\textwidth]{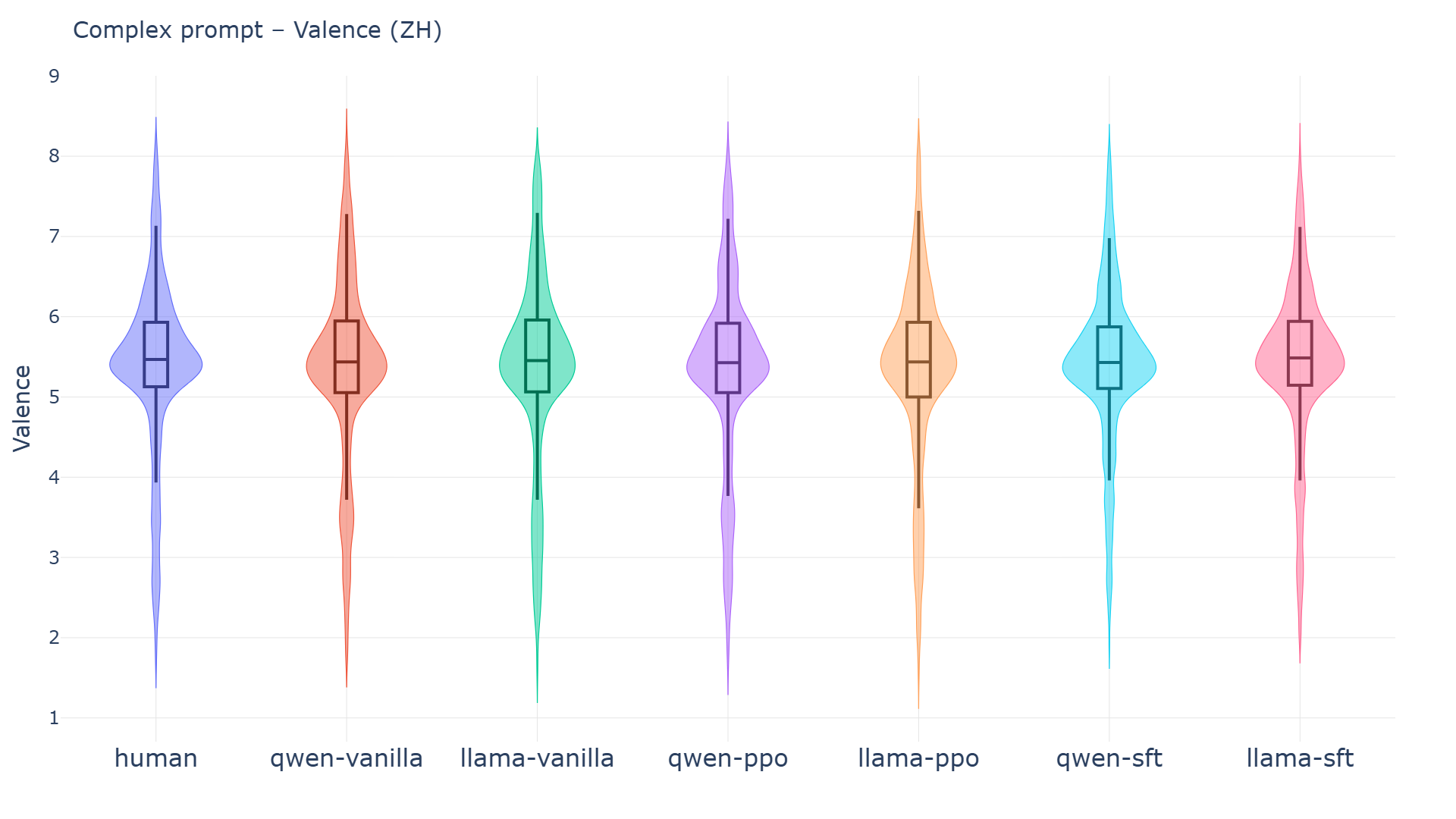}
        \caption{Mandarin: association \textbf{valence}, rescaled to the English 1–9 range.}
        \label{fig:valence_zh}
    \end{subfigure}
    \caption{Violin + box plots of per-cue \textbf{valence} medians for the \emph{Complex} prompt. Left: English (1 = unpleasant, 9 = pleasant); Right: Mandarin (rescaled to 1–9).}
    \label{fig:violin_plot_valence}
\end{figure}

\begin{figure}[!t]
\centering
    \captionsetup{font=small}
    \begin{subfigure}{0.46\textwidth}
        \centering
        \includegraphics[width=\textwidth]{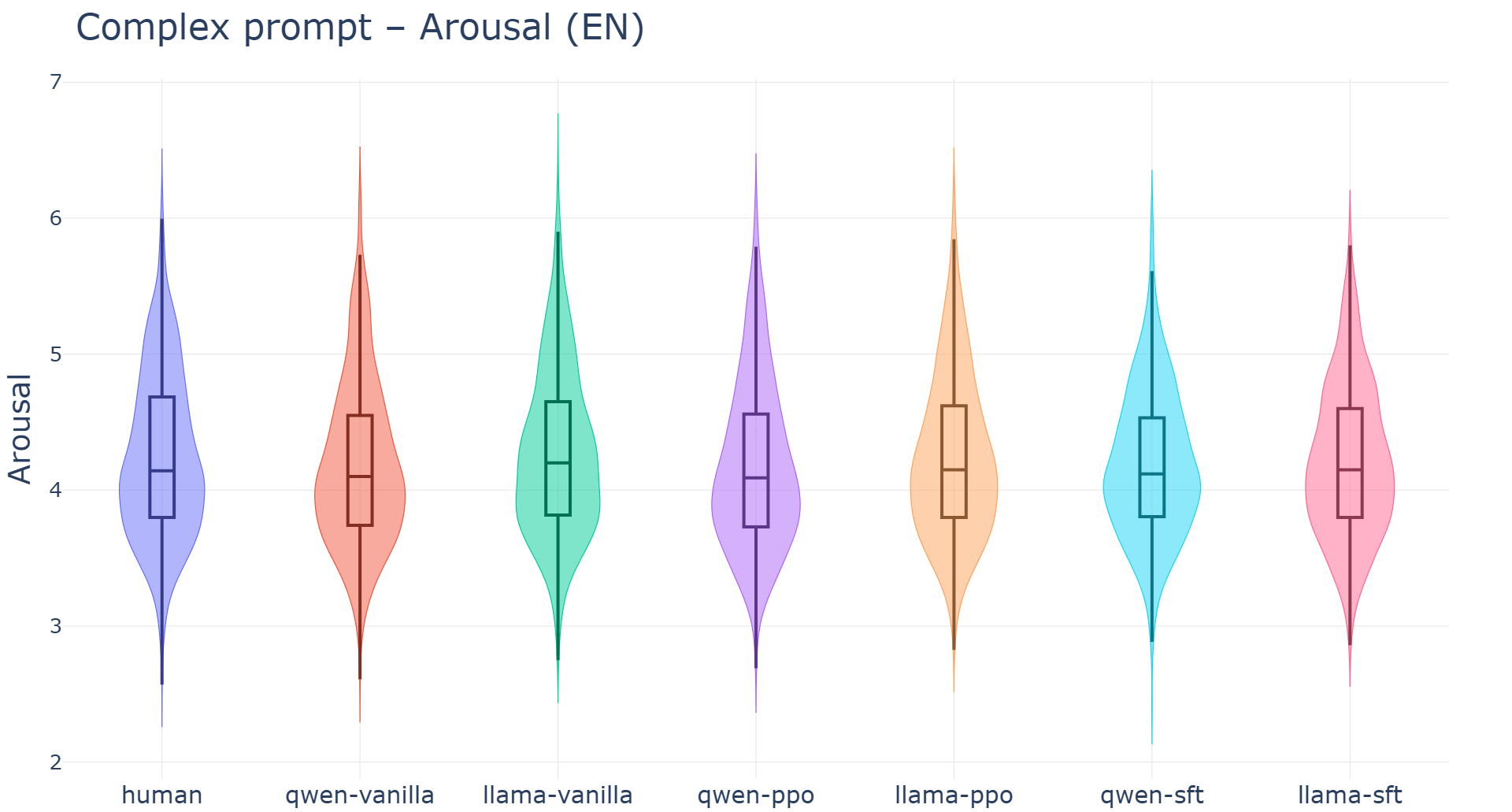}
        \caption{English: association \textbf{arousal} (1 = calm, 9 = excited).}
        \label{fig:arousal_en}
    \end{subfigure}
    \hfill
    \begin{subfigure}{0.46\textwidth}
        \centering
        \includegraphics[width=\textwidth]{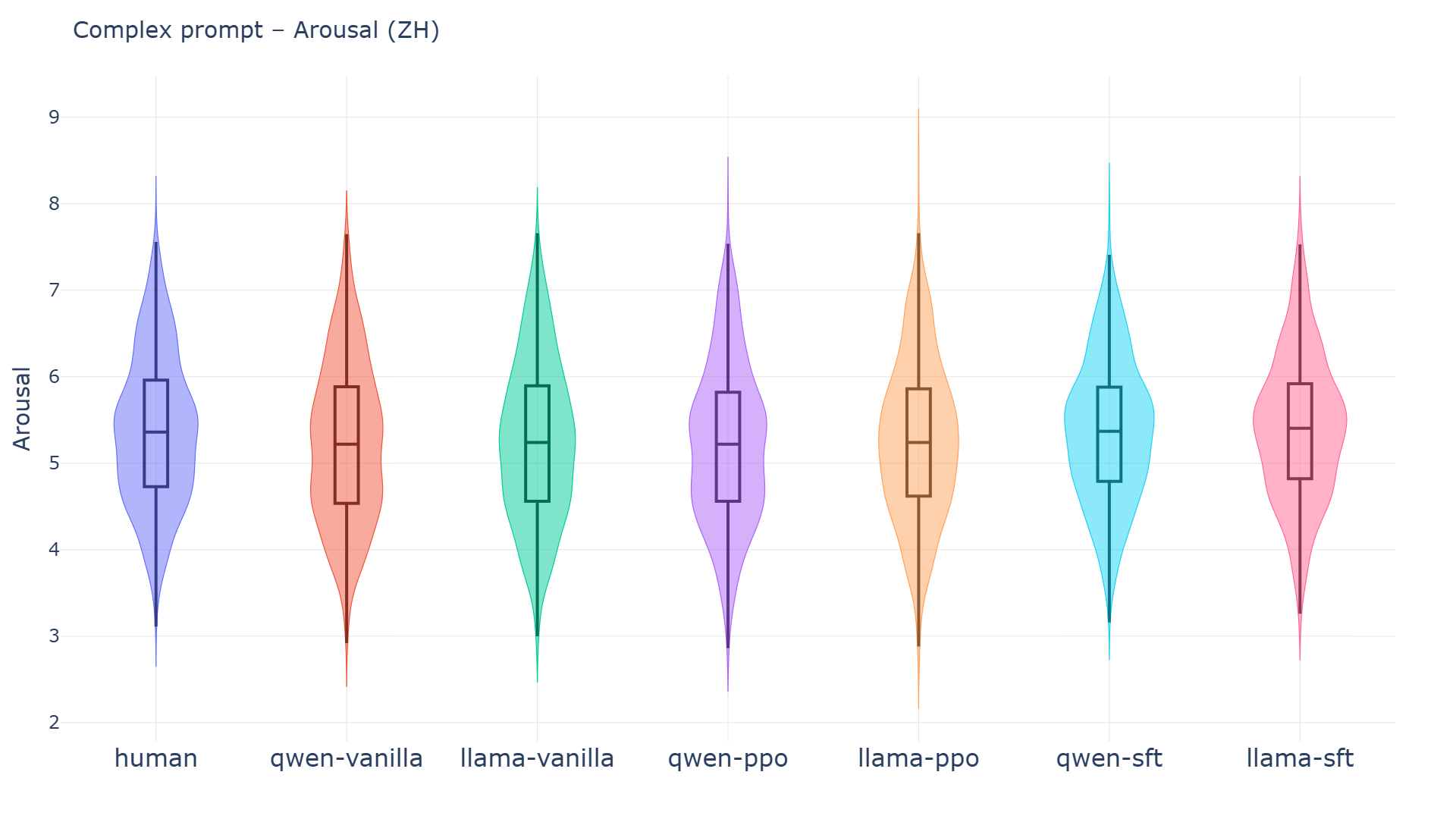}
        \caption{Mandarin: association \textbf{arousal}, rescaled to the English 1–9 range.}
        \label{fig:arousal_zh}
    \end{subfigure}
    \caption{Violin + box plots of per-cue \textbf{arousal} medians for the \emph{Complex} prompt. Left: English (1 = calm, 9 = excited); Right: Mandarin (rescaled to 1–9).}
    \label{fig:violin_plot_arousal}
\end{figure}

\subsection{Examples of psychological attributes}
\label{apdx:ssec:example_psychological_attributes}
Table~\ref{tab:examples_psychological_attributes} presents examples of the various sources of associations for two cues: Halloween and emotions. 
Below, we summarize key patterns observed in these examples.

\begin{table*}[!t]
\centering
\small 

\resizebox{\textwidth}{!}{%
\renewcommand{\arraystretch}{1.2}
\begin{tabular}{lllp{10cm}c}
\toprule
\textbf{Cue} & \textbf{Source} & \textbf{Attribute} & \textbf{Associations (Score)} & \textbf{Median} \\
\midrule
\multirow{12}{*}{\rotatebox{90}{\textbf{Halloween}}}
& Human & Valence & candy (7.27), pumpkin (7.0), costume (6.05), costumes (6.05), holiday (7.18), October (-), ghosts (4.23), orange (6.81), pumpkins (7.0), party (7.18) & 7.00 \\
& Llama-8B (Vanilla) & Valence & pumpkin (7.0), costume (6.05), trick-or-treat (5.87), candy (7.27), ghost (4.23), monster (2.55), skeleton (4.37), bat (4.81), spider (3.35), witch (3.14) & 4.59 \\
& Llama-8B (SFT) & Valence & costume (6.05), october (-), night (6.68), leyton (-), pumpkins (7.0), masks (4.81), holiday (7.18), trick-or-treat (5.87), scare (3.55), kid (7.23) & 6.37 \\
& Llama-8B (PPO) & Valence & costume (6.05), pumpkin (7.0), candy (7.27), trick-or-treat (5.87), ghost (4.23), spider (3.35), witch (3.14), candy corn (6.61), bats (4.81), black cat (6.18) & 5.96 \\ 
\cmidrule{2-5}
& Human & Arousal & candy (5.03), pumpkin (3.43), costume (4.78), costumes (4.78), holiday (4.93), October (-), ghosts (5.7), orange (4.04), pumpkins (3.43), party (6.08) & 4.78 \\
& Llama-8B (Vanilla) & Arousal & pumpkin (3.43), costume (4.78), trick-or-treat (5.29), candy (5.03), ghost (5.7), monster (5.55), skeleton (4.55), bat (4.57), spider (6.91), witch (5.3) & 5.16 \\
& Llama-8B (SFT) & Arousal & costume (4.78), october (-), night (3.57), leyton (-), pumpkins (3.43), masks (3.26), holiday (4.93), trick-or-treat (5.29), scare (7.1), kid (4.71) & 4.76 \\
& Llama-8B (PPO) & Arousal & costume (4.78), pumpkin (3.43), candy (5.03), trick-or-treat (5.29), ghost (5.7), spider (6.91), witch (5.3), candy corn (4.23), bats (4.57), black cat (4.04) & 4.91 \\ 
\cmidrule{2-5}
& Human & Concreteness & candy (4.83), pumpkin (4.9), costume (4.57), costumes (4.57), holiday (2.86), October (2.81), ghosts (3.19), orange (4.66), pumpkins (4.9), party (3.89) & 4.57 \\
& Llama-8B (Vanilla) & Concreteness & pumpkin (4.9), costume (4.57), trick-or-treat (3.36), candy (4.83), ghost (3.19), monster (3.72), skeleton (4.97), bat (5.0), spider (4.97), witch (4.17) & 4.70 \\
& Llama-8B (SFT) & Concreteness & costume (4.57), october (2.81), night (4.52), leyton (-), pumpkins (4.9), masks (4.96), holiday (2.86), trick-or-treat (3.36), scare (2.96), kid (4.56) & 4.52 \\
& Llama-8B (PPO) & Concreteness & costume (4.57), pumpkin (4.9), candy (4.83), trick-or-treat (3.36), ghost (3.19), spider (4.97), witch (4.17), candy corn (4.9), bats (5.0), black cat (4.31) & 4.70 \\
\midrule

\multirow{12}{*}{\rotatebox{90}{\textbf{emotions}}}
& Human & Valence & feelings (6.5), sad (2.1), happy (8.47), love (8.0), sadness (2.4), anger (2.5), angry (2.53), tears (3.14), cry (3.22), mad (2.47) & 2.83 \\
& Llama-8B (Vanilla) & Valence & feelings (6.5), sentiments (6.2), moods (5.29), sincerity (7.9), empathy (7.29), compassion (7.9), love (8.0), anger (2.5), fear (2.93), joy (8.21) & 6.90 \\
& Llama-8B (SFT) & Valence & feelings (6.5), happy (8.47), sad (2.1), love (8.0), mood (5.29), feelings and thoughts (6.63), heart (6.95), thoughts (6.76), feelings* (6.5), face (6.36) & 6.56 \\
& Llama-8B (PPO) & Valence & happiness (8.48), sadness (2.4), anxiety (2.38), love (8.0), fear (2.93), anger (2.5), empathy (7.29), jealousy (2.58), guilt (2.29), shame (2.62) & 2.60 \\ 
\cmidrule{2-5}
& Human & Arousal & feelings (3.86), sad (3.49), happy (6.05), love (5.36), sadness (2.81), anger (5.93), angry (6.2), tears (4.8), cry (5.45), mad (5.59) & 5.41 \\
& Llama-8B (Vanilla) & Arousal & feelings (3.86), sentiments (3.54), moods (4.5), sincerity (4.42), empathy (3.62), compassion (4.5), love (5.36), anger (5.93), fear (6.14), joy (5.55) & 4.50 \\
& Llama-8B (SFT) & Arousal & feelings (3.86), happy (6.05), sad (3.49), love (5.36), mood (4.5), feelings and thoughts (4.01), heart (5.07), thoughts (4.16), feelings* (3.86), face (4.59) & 4.33 \\
& Llama-8B (PPO) & Arousal & happiness (6.5), sadness (2.81), anxiety (4.78), love (5.36), fear (6.14), anger (5.93), empathy (3.62), jealousy (5.45), guilt (4.48), shame (5.4) & 5.38 \\ 
\cmidrule{2-5}
& Human & Concreteness & feelings (1.68), sad (3.07), happy (2.56), love (2.07), sadness (1.82), anger (2.41), angry (2.53), tears (4.56), cry (4.0), mad (2.76) & 2.54 \\
& Llama-8B (Vanilla) & Concreteness & feelings (1.68), sentiments (2.1), moods (1.75), sincerity (1.97), empathy (1.63), compassion (1.89), love (2.07), anger (2.41), fear (2.57), joy (2.37) & 2.02 \\
& Llama-8B (SFT) & Concreteness & feelings (1.68), happy (2.56), sad (3.07), love (2.07), mood (1.75), feelings and thoughts (1.68), heart (4.52), thoughts (1.97), feelings* (1.68), face (4.87) & 2.02 \\
& Llama-8B (PPO) & Concreteness & happiness (2.6), sadness (1.82), anxiety (2.21), love (2.07), fear (2.57), anger (2.41), empathy (1.63), jealousy (1.8), guilt (1.93), shame (2.24) & 2.14 \\
\bottomrule
\end{tabular}
}
\caption{Examples of cue–association psychological attributes, including Valence, Arousal, and Concreteness, for the cues \textit{Halloween} and \textit{emotions}. The ``Median'' column is the median score of each cue, computed from the attribute values of its associated words in each row. Numbers in parentheses indicate the corresponding attribute scores. Valence and Arousal values range from 1–9 (higher values indicate more pleasantness and stronger emotional intensity), while Concreteness values range from 1 (more abstract) to 5 (higher concrete). (-) means the word is not found in the corresponding norms.}
\label{tab:examples_psychological_attributes}
\end{table*}

\paragraph{Analysis on Valence.}
Fine-tuned models align their associations more closely with human representations, as reflected in the valence scores of their associations. 
For instance, when prompted with \textit{Halloween}, US participants tend to produce highly pleasant associations (median valence = 7), such as \textit{candy} (7.27), \textit{holiday} (7.18), and \textit{party} (7.18). 
In contrast, Vanilla Llama model often evokes less pleasant associations, including \textit{monster} (2.55), \textit{skeleton} (4.37), and \textit{spider} (3.35). 
Models fine-tuned on \swus narrow this gap: the Llama SFT model, for example, generates high-valence associations like \textit{holiday} (7.18) and \textit{kid} (7.23), more closely mirroring human affective patterns.
A similar pattern is observed for the cue \textit{emotions}. A substantial valence gap exists between human associations (median = 2.83) and those of the vanilla Llama model (6.90). The vanilla model tends to produce overly pleasant (high Valence scores) associations, whereas human associations reflect a more complex emotional landscape—mixing both positive and negative feelings such as \textit{sad} (2.1), \textit{happy} (8.47), \textit{anger} (2.5), and \textit{cry} (3.22). The best-performing fine-tuned model, Llama-PPO, better captures this complexity, generating associations such as \textit{happiness} (8.48), \textit{sadness} and \textit{anxiety} (2.38).

\paragraph{Analysis on Arousal.}
Unlike valence, which captures pleasantness, arousal reflects the intensity or activation level of associations. For \textit{Halloween}, human responses indicate moderate excitement (median = 4.78), balancing calm elements like \textit{pumpkin} (3.43) with livelier cues such as \textit{party} (6.08). The vanilla models amplify this excitement, favoring highly stimulating words like \textit{monster} (5.55) and \textit{spider} (6.91), resulting in slightly higher overall arousal (median = 5.16). Fine-tuned variants temper this tendency—Llama-PPO (median = 4.91), for instance, retrieves a steadier mix of associations spanning both neutral and intense states (\textit{candy} (5.03), \textit{ghost} (5.7), \textit{black cat} (4.04)).

The cue \textit{emotions} shows an example that the Vanilla model tends to diminish the emotional intensity. Human associations cover a broad emotional range, from \textit{sad} (3.49) to \textit{anger} (5.93), yielding a balanced median (5.41). Vanilla Llama compresses this variation, producing a flatter, less expressive pattern (median = 4.50). In contrast, the Llama-PPO model restores much of this dynamic spread (median = 5.38), surfacing high-arousal concepts such as \textit{anger} (5.93), \textit{guilt} (4.48), and \textit{shame} (5.40), which better approximate human affective diversity.

\paragraph{Analysis on Concreteness.}
As observed previously, models still struggle to align their concreteness scores with human judgments. For \textit{Halloween}, the Llama SFT model achieves a similar median concreteness score to humans, whereas other models produce overly concrete associations. The pattern reverses for the cue \textit{emotions}, where models tend to generate words that are excessively abstract.

\section{Evaluation Results on World Values Survey}
\label{apex:eval_WVS}
\subsection{Breakdown results on EMD}
\label{apdx:ssec:percentage_emd}
\begin{figure}[!t]
    \centering
        
    
    \begin{subfigure}{0.45\textwidth}
        \centering 
        \includegraphics[width=\textwidth]{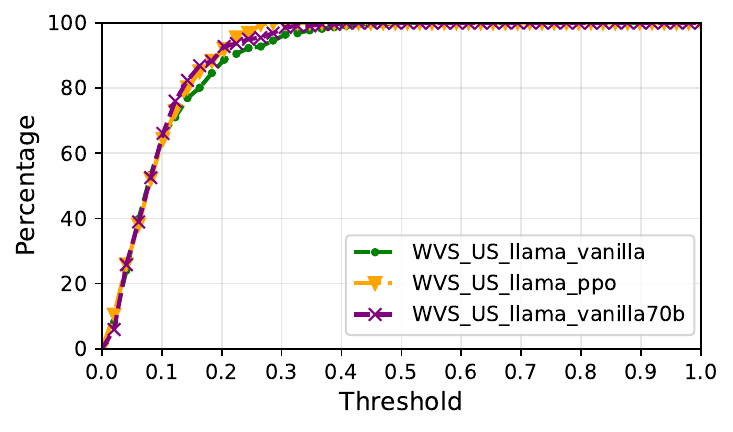}
        \caption{WVS-us under Jensen Shnnon}
        \label{fig:WVS_us_under_Jensen_Shnnon}
    \end{subfigure}
    \vspace{1pt}
    \begin{subfigure}{0.45\textwidth}
        \centering
        \includegraphics[width=\textwidth]{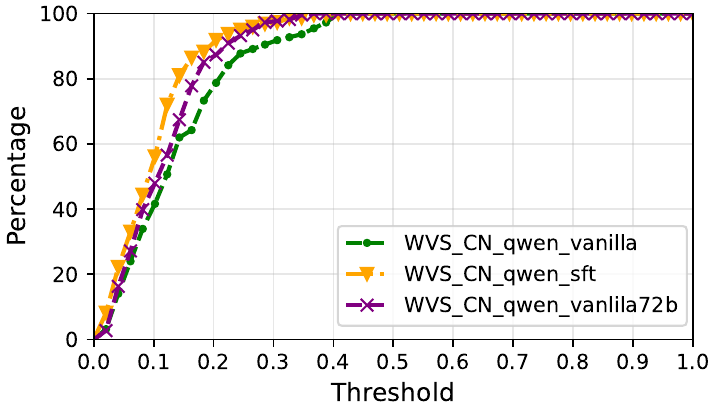}
        \caption{WVS-zh performance under Jensen Shannon }
        \label{fig:WVS_zh_performance_under_Jensen_Shannon}
    \end{subfigure}
    \caption{Breakdown comparison of model alignment with cultural values across China and  United States based on the World Values Survey. Results are shown for the Vanilla and trained (SFT and PPO) versions of Qwen2.5 and Llama 3.1.}
    \label{fig:wvs_results_plots_emd}
\end{figure}

\subsection{Tension Set Selection}
\label{apdx:ssec:tension_set}
Given the participants' answer distributions for China $(q)$ and the United States $(p)$, we first normalise each to a probability vector i.e. we divide each count by the total number of respondents for that question so the values now represent probabilities (fractions between 0 and 1).  Divergence is then measured with a hybrid score that averages an entropy-sensitive component (Jensen–Shannon divergence, $JSD$) and an ordinal component (normalised Earth-Mover distance, $EMD$):
\[
\mathrm{combo}(p,q)
  \;=\;
  \tfrac{1}{2}\,JSD(p,q)
  \;+\;
  \tfrac{1}{2}\,EMD(p,q).
\]

Sorting the WVS questions by this score and retaining the
top~$50$ yields our fixed \textit{tension set}.

\subsection{Cross-Cultural Value Alignment Evaluation (EN Prompts)}
\label{appendix:ssec:en_prompt_alignment}

\begin{figure}[!t]
  \centering
  \begin{subfigure}[b]{\columnwidth}
    \centering
    \includegraphics[width=\columnwidth, height=0.3\textheight, keepaspectratio]{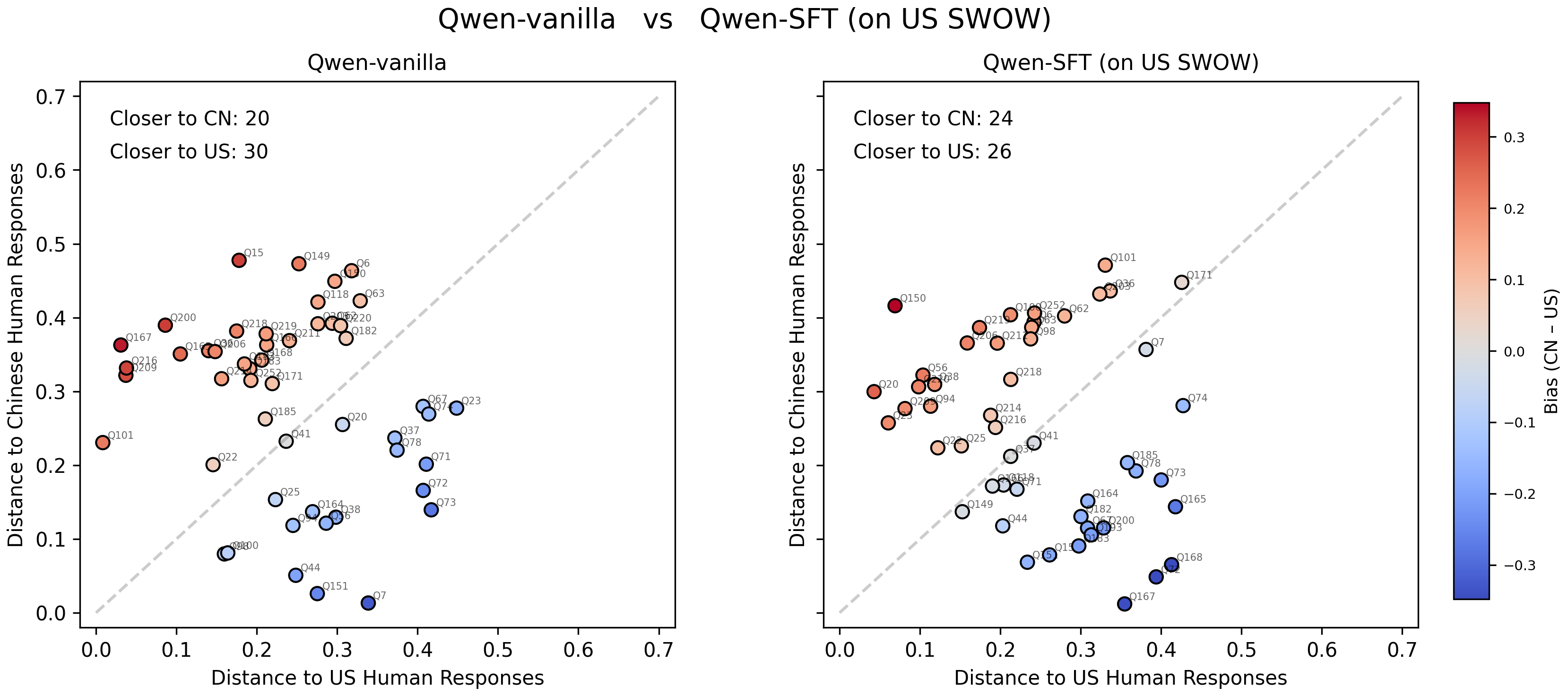}
    \caption{\textbf{Qwen-7B}: SFT on US SWOW does not shift the cloud substantially.}
    \label{fig:appendix_qwen_en}
  \end{subfigure}
  
  \vspace{1em}
  
  \begin{subfigure}[b]{\columnwidth}
    \centering
    \includegraphics[width=\columnwidth, height=0.3\textheight, keepaspectratio]{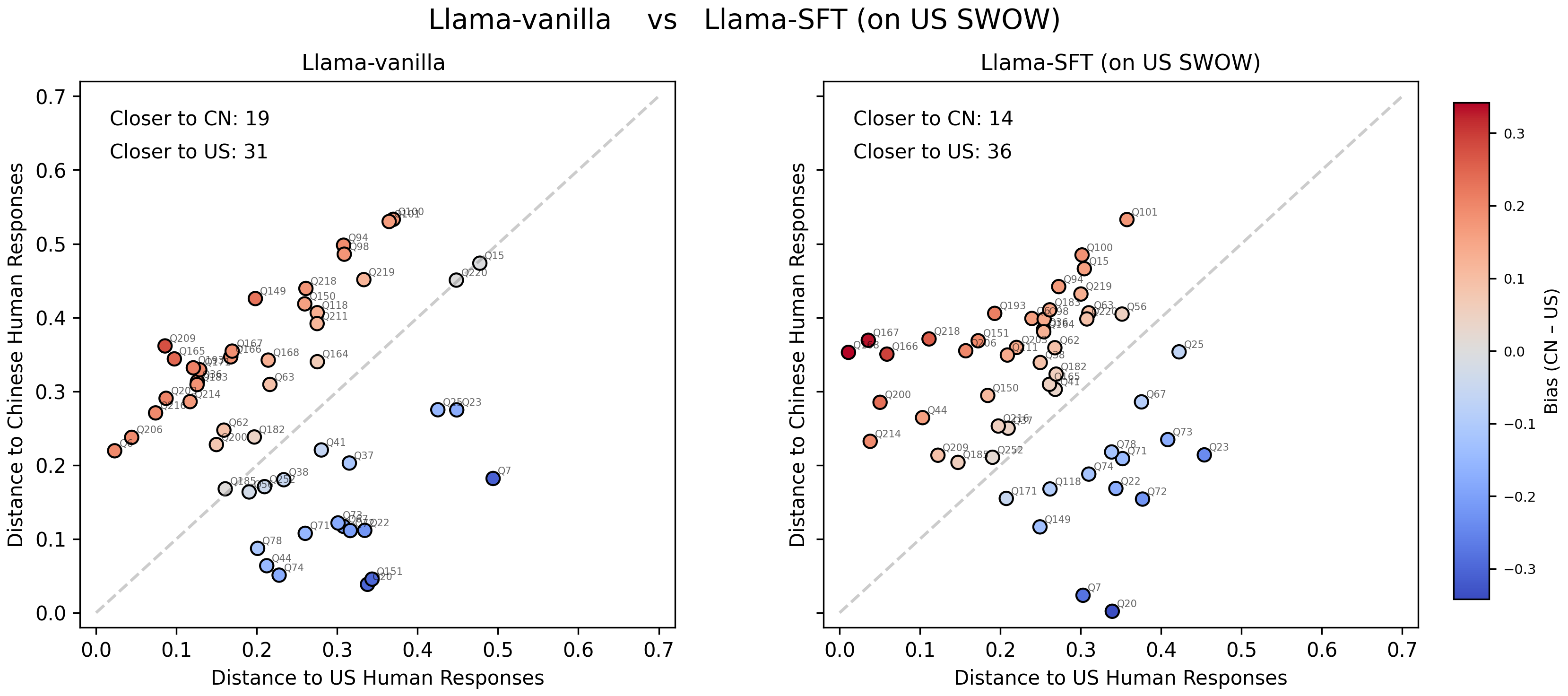}
    \caption{\textbf{Llama-8B}: minimal movement after SFT on US SWOW.}
    \label{fig:appendix_llama_en}
  \end{subfigure}
  
  \caption{Shifts after SFT on US SWOW (EN prompts). Each dot = one WVS question; colour = bias (CN–US).}
  \label{fig:appendix_shift_comparison_en}
\end{figure}

Beyond Mandarin prompts, we also evaluate cultural shifts with English prompts.
Figures~\ref{fig:appendix_qwen_en} and~\ref{fig:appendix_llama_en}
mirror the same layout used for Chinese prompts: hybrid distances to
US answers (x-axis) and Chinese answers (y-axis) are plotted across 50 high-tension WVS questions.

\begin{CJK*}{UTF8}{gbsn}
\begin{table*}[!t]
  {%
  \centering
  \rowcolors{2}{gray!15}{white}
  \resizebox{\textwidth}{!}{%
    \begin{tabular}{p{2.5cm} p{4cm} p{1.5cm} c c c c c c c c}
      \toprule
      \rowcolor{gray!30}
      \textbf{Question (ZH)} & \textbf{Prompt (EN)} & \textbf{Survey} &
      \textbf{Q\textsubscript{van}} & \cellcolor{blue!10}\textbf{Q\textsubscript{sft}} &
      \textbf{JS} & \cellcolor{blue!10}\textbf{JS–SFT} &
      \textbf{EMD} & \cellcolor{blue!10}\textbf{EMD–SFT} &
      \textbf{Type} \\
      \midrule
      您是否认为有天堂？ &
      In which of the following do you believe, if you believe in any? – Heaven {\itshape(1: Yes; 2: No)} &
      [12\%,88\%] &
      [71\%,29\%] & \cellcolor{blue!5}[18\%,82\%] &
      0.437 & \cellcolor{blue!5}0.061 &
      0.173 & \cellcolor{blue!5}0.062 &
      Religious \\

      您是否相信死后有来生？ &
      In which of the following do you believe, if you believe in any? – Life after death {\itshape(1: Yes; 2: No)} &
      [12\%,88\%] &
      [90\%,10\%] & \cellcolor{blue!5}[36\%,64\%] &
      0.596 & \cellcolor{blue!5}0.208 &
      0.020 & \cellcolor{blue!5}0.246 &
      Religious \\

      您是否信仰佛祖/上帝/真主/神明？ &
      In which of the following do you believe, if you believe in any? – God {\itshape(1: Yes; 2: No)} &
      [17\%,83\%] &
      [41\%,59\%] & \cellcolor{blue!5}[29\%,71\%] &
      0.182 & \cellcolor{blue!5}0.100 &
      0.232 & \cellcolor{blue!5}0.119 &
      Religious \\

      您是否认为有地狱？ &
      In which of the following do you believe, if you believe in any? – Hell {\itshape(1: Yes; 2: No)} &
      [11\%,89\%] &
      [47\%,53\%] & \cellcolor{blue!5}[16\%,84\%] &
      0.288 & \cellcolor{blue!5}0.049 &
      0.359 & \cellcolor{blue!5}0.047 &
      Religious \\
      \bottomrule
    \end{tabular}%
  }
  \caption{Comparison of survey distributions and model outputs (vanilla vs.\ SFT) for five religious-belief WVS items. Highlighted cells show metrics after SFT.}
  \label{tab:example_distribution_shift}
  }
\end{table*}
\end{CJK*}

\begin{itemize}[leftmargin=*]
  \item \textbf{Qwen-7B.}  
    The vanilla model already exhibits strong alignment with US responses; fine-tuning on US SWOW slightly reduces this alignment (from 30 to 26 US-aligned points).
  
  \item \textbf{Llama-8B.}  
    Supervised fine-tuning increases US alignment, shifting the number of US-aligned points from 31 to 36.
\end{itemize}

These results suggest that for English prompts, vanilla models—particularly Qwen—may already exhibit strong US alignment, reducing the effect of SFT on US SWOW.

\subsection{WVS Answer Shifts Across Topics}
\label{app:wvs_topic_shifts}

To examine fine-grained cultural effects, we group WVS questions into twelve topical domains and compare alignment before and after SFT on Chinese SWOW.  
Figures~\ref{fig:WVS_ZH_vanilla_SFT_JS_topics} and~\ref{fig:WVS_ZH_Vanillia_SFT_EMD_topics} (below) visualize Jensen–Shannon and Earth Mover’s distances by topic.  
Fine-tuning improves alignment in five domains—ethical values, political engagement, religious beliefs, social capital, and safety perceptions—while it slightly reduces alignment for economic values and corruption perceptions.  
This drop may reflect a mismatch between model training distributions and the nuanced economic attitudes Chinese respondents hold.

\begin{figure}[!h]
  \centering
  \includegraphics[width=0.8\linewidth,height=0.7\textheight,keepaspectratio]{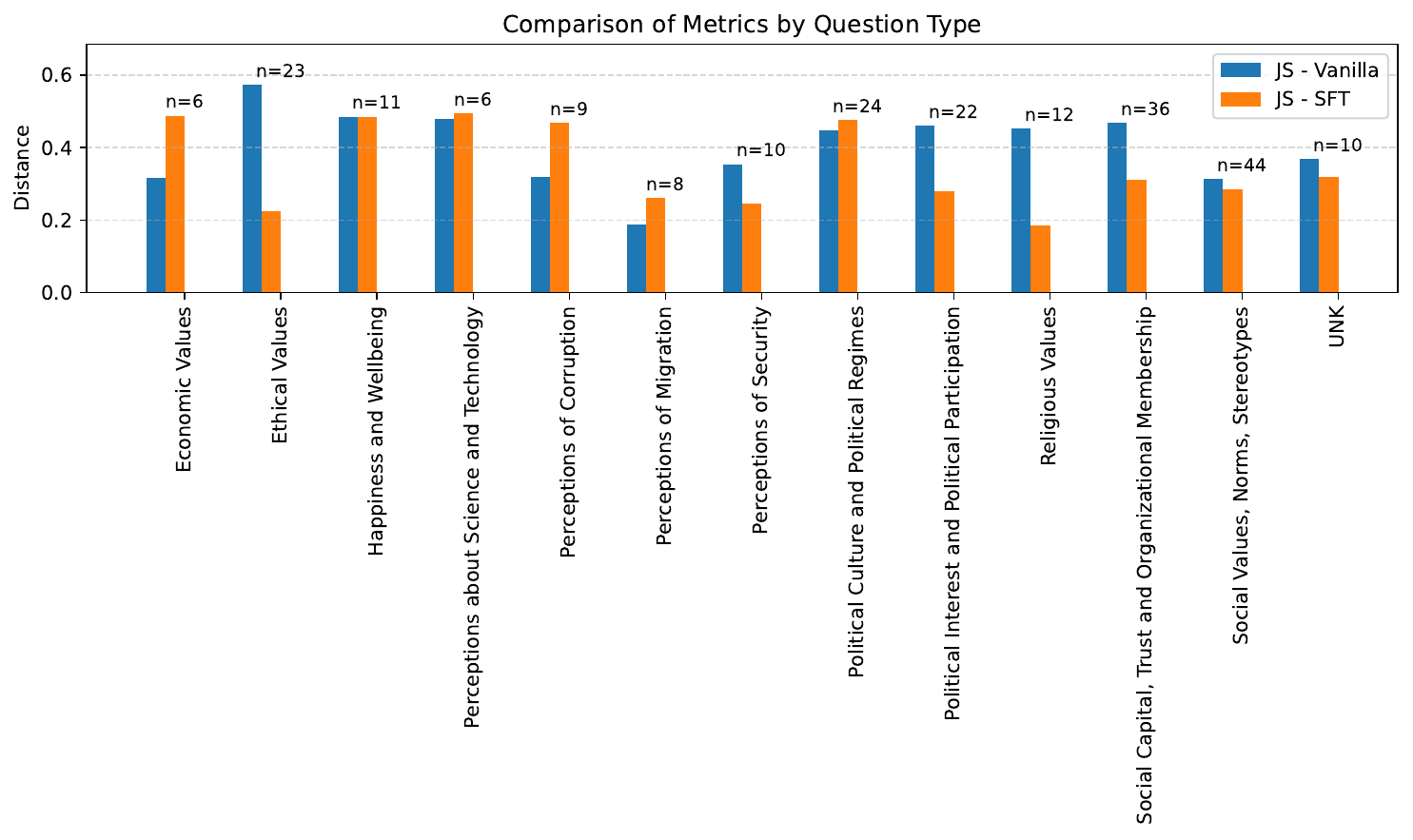}
  \caption{Jensen–Shannon distance by WVS topic (Vanilla vs.\ SFT Qwen-7B on ZH prompts).}
  \label{fig:WVS_ZH_vanilla_SFT_JS_topics}
\end{figure}

\begin{figure}[!h]
  \centering
  \includegraphics[width=0.8\linewidth,height=0.7\textheight,keepaspectratio]{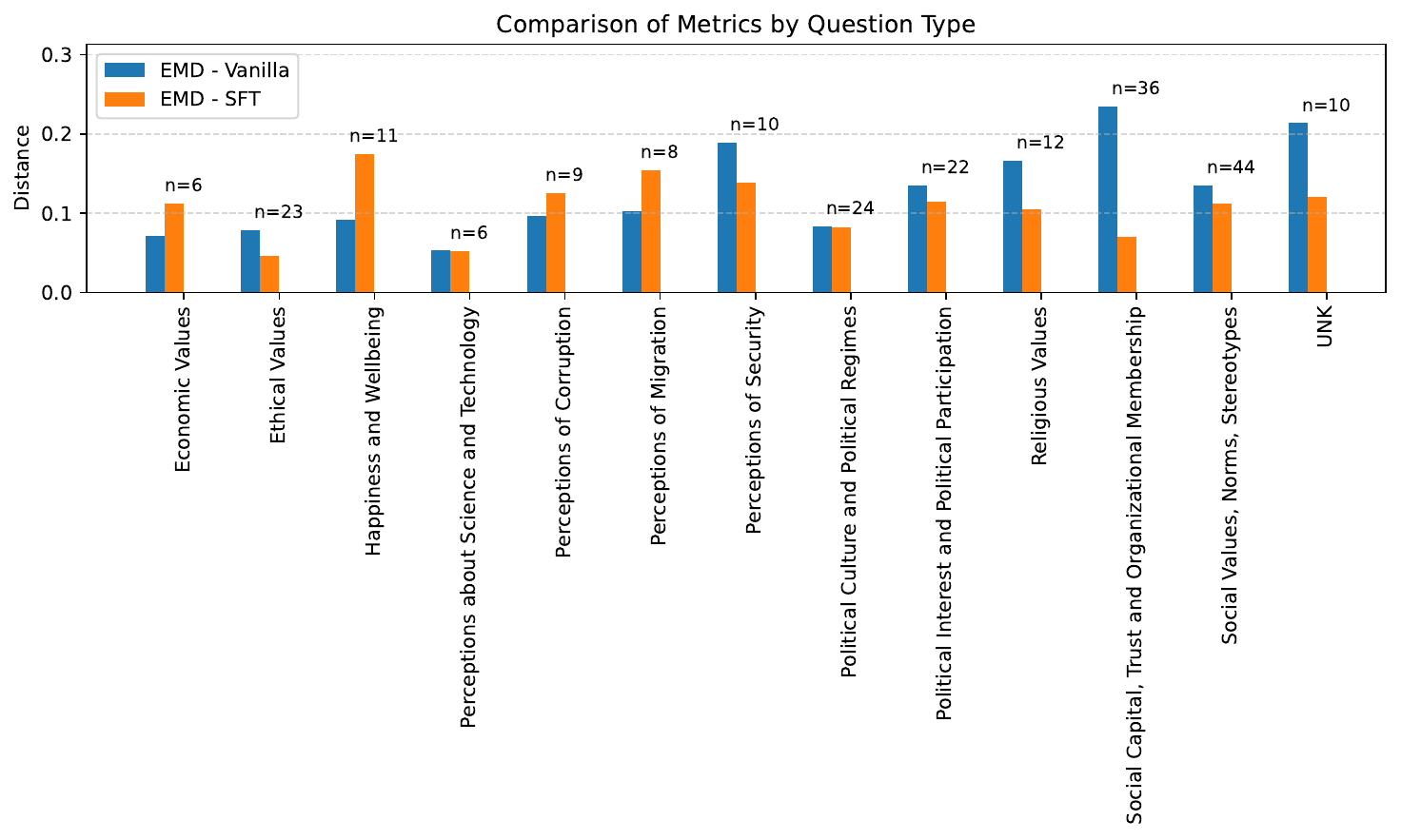}
  \caption{Earth Mover’s distance by WVS topic (Vanilla vs.\ SFT Qwen-7B on ZH prompts).}
  \label{fig:WVS_ZH_Vanillia_SFT_EMD_topics}
\end{figure}

\begin{CJK*}{UTF8}{gbsn}
Table~\ref{tab:example_distribution_shift} presents concrete examples of distribution shifts from the vanilla Qwen-2.5 model to the SFT Qwen-2.5 model. For example, in the domain of religious values, the vanilla model’s predictions are either overly dispersed or peak at culturally incongruent options, whereas fine-tuning realigns the predicted distributions with human responses.  
When asked “Do you believe in Heaven?”, the vanilla model strongly predicts “Yes” (0.70), while the fine-tuned model shifts to “No” (0.84), closely matching the actual distribution from Chinese participants (0.89 “No”).  
Notably, although the SFT model rejects Western religious imagery like “Heaven,” it also captures Chinese-specific spiritual concepts such as “Life after death.”  
In the SWOW–ZH associations for \textit{死亡} (death), responses like \textit{轮回} (reincarnation) and \textit{新生} (new life) reflect how Chinese speakers conceptualize death, illustrating how association-based fine-tuning contributes to value prediction.
\end{CJK*}

\end{document}